\pdfoutput=1
\documentclass[11pt]{article}

\usepackage{acl}

\usepackage{times}
\usepackage{latexsym}

\usepackage[T1]{fontenc}
\usepackage[utf8]{inputenc}

\usepackage{microtype}

\usepackage{inconsolata}

\usepackage{amsmath, amssymb}
\usepackage{makecell}
\usepackage{amsfonts}  % some math symbol  for example, /mathbb{}
\usepackage{graphicx}  % resize tables and figures
\usepackage{boldline}  % for table
\usepackage{multirow}  % for table
\usepackage{booktabs}  % for table
\usepackage{array}     % for table
\usepackage{xcolor}    % for table
\usepackage{arydshln}
\usepackage{makecell}
\usepackage{floatrow}  % resize table for half page.
\usepackage{subcaption} % for figures
\usepackage[normalem]{ulem} % 提供 \sout

\newcommand\blfootnote[1]{%
  \begingroup
  \renewcommand\thefootnote{}\footnote{#1}%
  \addtocounter{footnote}{-1}%
  \endgroup
}

\title{Memorization, Emergence, and Explaining Reversal Failures: A Controlled Study of Relational Semantics in LLMs}

\makeatletter
\renewcommand{\thefootnote}{\fnsymbol{footnote}}
\makeatother

\author{
Yihua Zhu$^{1,3}$ \qquad  Qianying Liu$^{3}$\textsuperscript{\dag} \qquad  Jiaxin Wang$^{1}$ \qquad Fei Cheng$^{1}$ \qquad Chaoran Liu$^{3}$ \qquad \\
\textbf{Akiko Aizawa}$^{2,3}$ \qquad \textbf{Sadao Kurohashi}$^{1,3}$ \qquad \textbf{Hidetoshi Shimodaira}$^{1,4}$
 \\
$^1$Kyoto University \qquad $^2$University of Tokyo \qquad $^3$NII LLMC \qquad $^4$RIKEN\\
\texttt{\{zhu.yihua.22h, wang.jiaxin.77y\}@st.kyoto-u.ac.jp}\\ 
\texttt{\{feicheng, kuro, shimo\}@i.kyoto-u.ac.jp}\\ 
\texttt{\{ying, cliu, aizawa\}@nii.ac.jp}}

\date{}

\begin{document}

\maketitle

\begingroup
\renewcommand\thefootnote{}\footnotetext{\textsuperscript{\dag} Corresponding author.}
\endgroup

%\footnotetext[2]{Corresponding author}

%................................abstract
\begin{abstract}

Autoregressive LLMs perform well on relational tasks that require linking entities via relational words (e.g., father/son, friend), but it is unclear whether they learn the logical semantics of such relations (e.g., symmetry and inversion logic) and, if so, whether reversal-type failures arise from missing relational semantics or left-to-right order bias. We propose a controlled Knowledge Graph-based synthetic framework that generates text from symmetric/inverse triples, trains GPT-style autoregressive models from scratch, and evaluates memorization, logical inference, and in-context generalization to unseen entities to address these questions. We find a sharp phase transition in which relational semantics emerge with sufficient logic-bearing supervision, even in shallow (2–3 layer) models, and that successful generalization aligns with stable intermediate-layer signals. Finally, order-matched forward/reverse tests and a diffusion baseline indicate that reversal failures are primarily driven by autoregressive order bias rather than deficient inversion semantics. 

\end{abstract}

\blfootnote{Our code is available at \url{https://github.com/YihuaZhu111/KG_synthetic_LLM_train}.}

\begin{figure}[!t]
%\centering
\includegraphics[width=1\columnwidth]{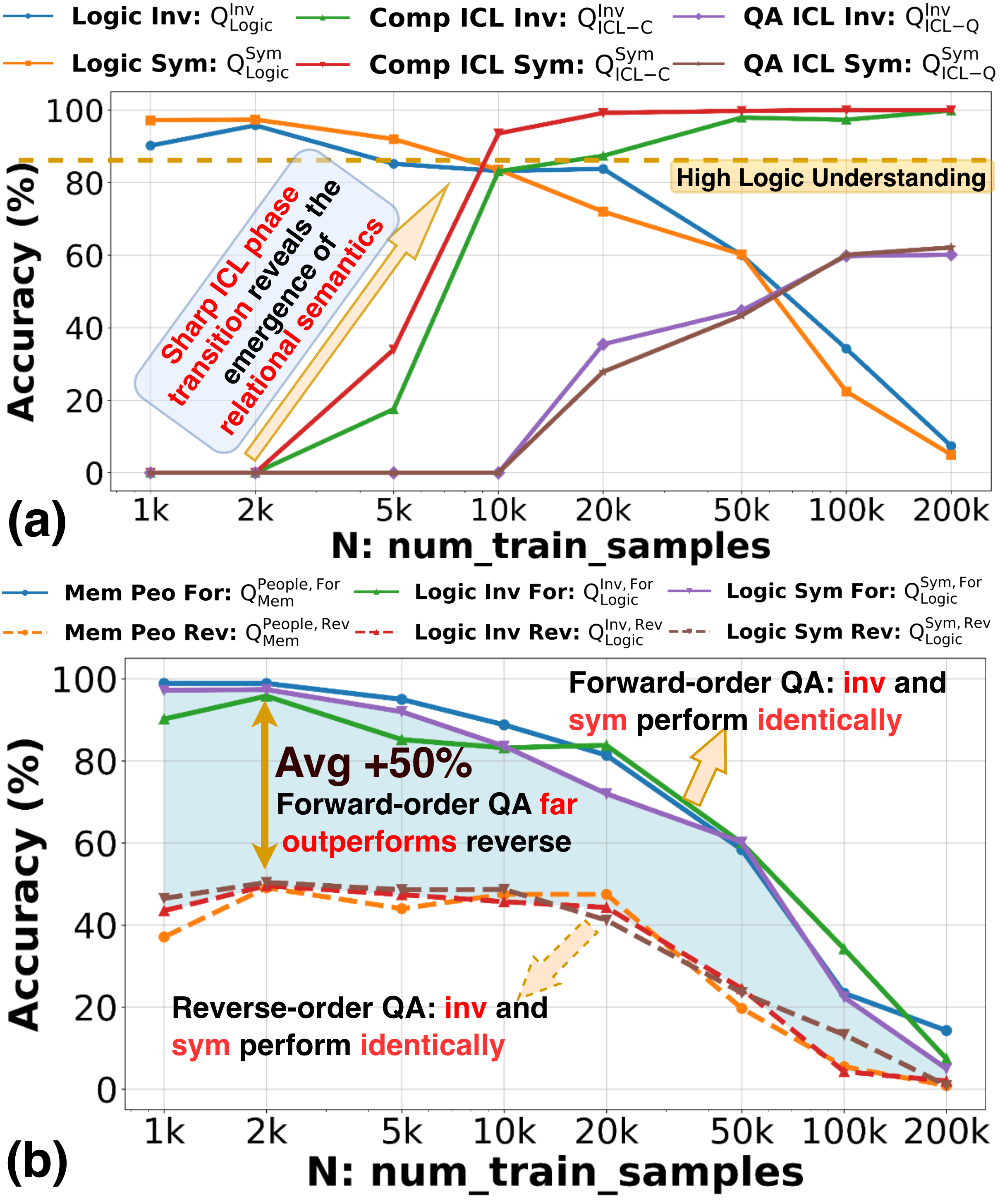}

\caption{\textbf{Two-stage inquiry into relational-word understanding in autoregressive language models.} (a) Models can memorize symmetric and inverse relational facts and perform forward-order logic QA; both completion- and QA-based in-context evaluations on unseen entities exhibit a sharp phase transition with sufficient logic-bearing supervision, indicating the emergence of relational semantics. (b) While inverse relations perform comparably to their original under forward queries, reversed-order queries show a substantial performance drop, revealing the reversal curse of order sensitivity.}
\label{fig:emergence_reverse}
\end{figure}

% Introduction
\section{Introduction} \label{sec:intro}

Auto-regressive (AR) large language models (LLMs) trained on massive corpora have achieved remarkable progress across a wide range of NLP tasks, including question answering, information extraction, and reasoning \cite{vaswani2017attention, wei2022chain, wei2022emergent, pan2024unifying}. Many of these tasks fundamentally rely on the ability to recognize, represent and manipulate relations between entities. Despite this success, it remains unclear whether LLMs truly internalize the \emph{logical semantics} carried by \emph{relational words}, or whether their performance primarily reflects learning surface-level co-occurrence patterns from text.

Relational words occupy a special role in natural language: beyond linking entities in a sentence, they often encode abstract and systematic logical properties. For instance, some relations are \emph{symmetric} (if $A$ is a \texttt{friend} of $B$, then $B$ is a \texttt{friend} of $A$), while others form \emph{inverse} pairs (if $A$ is the \texttt{father} of $B$, then $B$ is the \texttt{son} of $A$). In knowledge graphs (KGs), such regularities are formalized as relation properties and serve as a core organizing principle for relational reasoning \cite{sun2019rotate, zhu-shimodaira-2024-3d}. This motivates our first research question:
\textbf{RQ1: Can auto-regressive language models memorize relational facts, and internalize the logical semantics of relational words, and under what training conditions (e.g., data and model scale) does this ability emerge?}

While RQ1 concerns whether relational semantics can be learned at all,
RQ2 focuses on how such knowledge—when present—is deployed at inference time.
Recent work reports the reversal curse \cite{berglund2023reversal}: a model trained on relational statements of the form ``$A$ is the father of $B$'' often fails to answer the logically equivalent reversed query ``$B$ is the son of $A$.'' This raises a fundamental ambiguity: are such failures caused by missing explicit inversion relational representations, or by order bias induced by left-to-right AR decoding? This leads to our second research question:
\textbf{RQ2: When a model has internalized relational semantics, are reversal-type failures primarily due to deficiencies in relational semantics, or to order bias in AR generation?} 
%Complementary, RQ1 and RQ2 intersect at a central issue: in LLMs, whether relation logic, represented by relational words, is in a form that supports robust, direction-agnostic inference.
Taken together, RQ1 and RQ2 converge on a central question: in LLMs, is the logic of relations encoded by relational words represented in a way that supports robust, direction-agnostic inference?

Answering these questions using standard web-scale pretrained LLMs poses two key challenges. Relational words are often polysemous and context-dependent, and uncontrolled pretraining data introduces contamination risks, making it difficult to distinguish genuine logical generalization from memorization \cite{allen2023physics}.

To address these challenges, we propose a fully controlled experimental framework based on synthetic corpora constructed from knowledge-graph triples that explicitly encode relational properties such as symmetry and inversion. We verbalize each triple using a small set of templates and compose them into paragraph-level text with controlled linguistic variation, enabling systematic manipulation of training data scale. Using this corpus, we then train GPT-style AR models from scratch using next-token prediction, followed by question--answer style supervised fine-tuning.
This framework enables a systematic examination of relational understanding under multiple evaluation settings. To examine RQ1, under both cloze-style sentence completion and question answering, we test whether models can perform (i) Memorize QA: memorize observed relational facts and answer questions, (ii) Logic QA: infer missing symmetric or inverse facts given partial information, and (iii) In-context Learning (ICL) Logic QA: generalize relational knowledge to unseen entities. To study RQ2, we further perform order-matched forward and reverse evaluations, and compare AR models with diffusion-based language models that do not rely on left-to-right decoding.

Using our controlled framework, as shown in Figure \ref{fig:emergence_reverse}, our main findings can be summarized as follows:

\begin{enumerate}
\item \textbf{Relational fact learning under autoregressive training.} We show that autoregressive LMs can memorize relational facts during pretraining and answer both memorize and logic questions when queried in the same forward order seen during training, with memorization capacity increasing with model scale.

\item \textbf{Emergence of relational semantics and generalization.} 
    We find that relational semantics emerge with sufficient logic-bearing supervision, enabling generalization to unseen entities. This emergence exhibits a sharp phase transition from zero to near perfect as training data increases, and can occur even in small, shallow (2–3 layers) models.

\item \textbf{Layer-wise correlates of relational generalization.} 
    Through layer-wise analysis, we show that successful relational generalization is associated with stable, logic-relevant representations in intermediate layers, whereas unsuccessful models exhibit late-stage, unstable signals that degrade at the final layer.

\item \textbf{Revisiting the reversal curse.} 
    We provide evidence that reversal-type failures primarily stem from order bias in left-to-right autoregressive decoding rather than missing relational semantics, and show that this issue is mitigated by bidirectional data exposure during training. In contrast, diffusion-based LMs do not exhibit such behavior.

\end{enumerate}

\begin{figure*}[t]
\centering
\includegraphics[width=1\columnwidth]{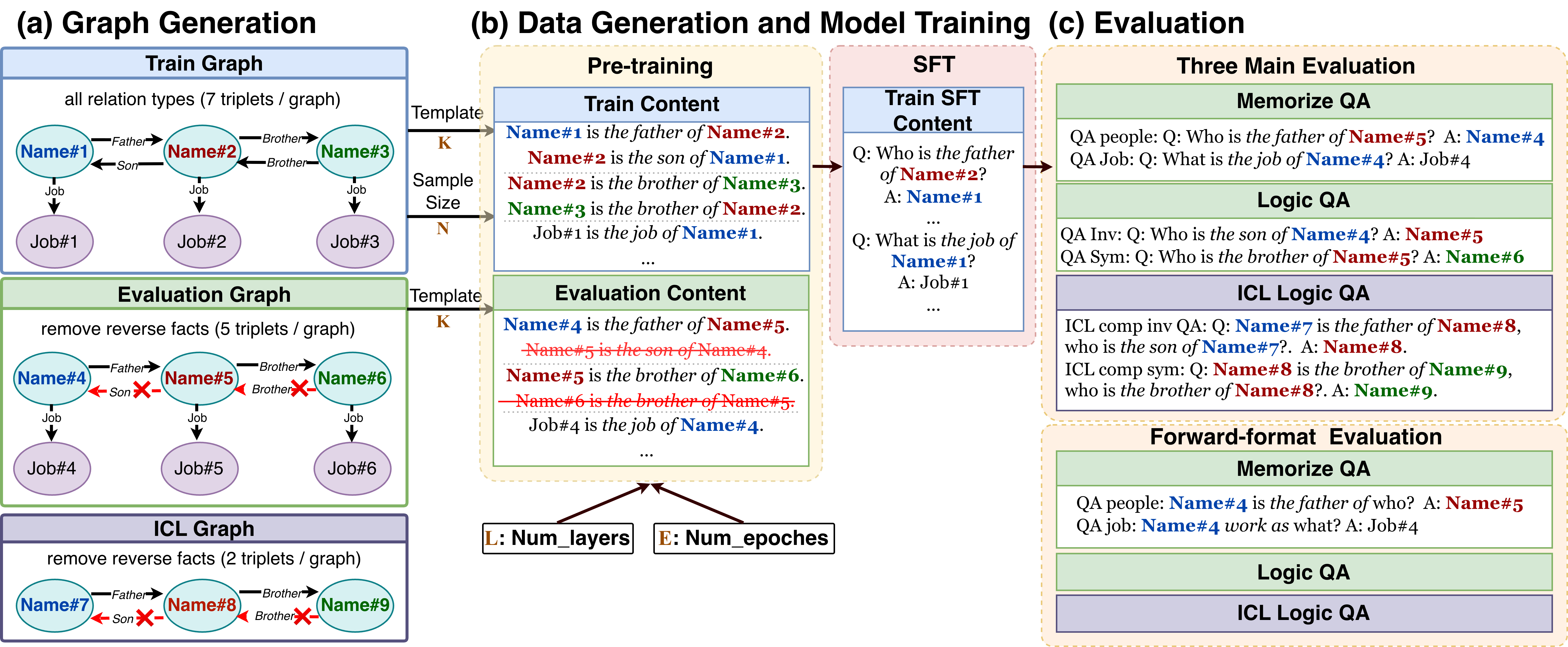}
\caption{Overview of the KG-synthetic data generation framework, model training process, and evaluation detail.}
\label{fig:Overall_framework}
\end{figure*}

\section{Methodology}\label{methodology}

This section defines our controlled experimental setup, as shown in Figure~\ref{fig:Overall_framework}.

\subsection{Controllable KG synthetic corpus}
\label{sec:prelim-synthetic}

We construct a fully controllable synthetic corpus grounded in a KG schema. We show an end-to-end text generation example in Table~\ref{tab:text_pipeline_examples}.

\paragraph{KG and triple generation.}

As shown in Figure~\ref{fig:Overall_framework}, each KG triple is generated by independently sampling its entities from uniform distributions. Person names take the form [first, middle, last], where each part is sampled from one of three disjoint pools of 100 synthetic tokens, yielding up to $10^6$ unique full names; job entities are sampled from 300 real-world occupations. Logic relations include (i) inversion kinship pairs (e.g., father/son, husband/wife, uncle/niece) and (ii) symmetric relations (e.g., brother, friend, spouse). The non-logic relation is (iii) a person–job relation. 

We construct three graph types: Train Graphs contain all relation types and include both directions for inversion and symmetric relations plus job facts (7 triples per graph); Evaluation Graphs remove the reverse facts (e.g., keep “name\#4 is the father of name\#5” but drop “name\#5 is the son of name\#4”); ICL Graphs keep only person–person relations and also remove reverse facts. Notably, names are sampled independently across the three graph types, so entity sets do not overlap.

\paragraph{Pre-train corpus construction.}

Triples are insufficient for training LLMs, so we verbalize each triple into a natural-language sentence by randomly choosing one of four surface formats (see Table~\ref{tab:templates}). For instance, $(\text{Name\#1}, \texttt{father}, \text{Name\#2})$ is rendered as ``Name\#1 is the father of Name\#2.'' For each KG, we sample $K$ distinct paragraph templates ($\texttt{num\_template}=K$) by selecting one format for each triple, which yields $4^{7}$ possible realizations for Train Graphs (7 triples) and $4^{5}$ for Evaluation Graphs (5 triples). We then randomly shuffle the sentence order within each paragraph. 

This yields $K$ templated paragraphs per graph. We then generate the train content from $N$ independently sampled Train Graphs (denoted as $\texttt{num\_train\_samples}=N$). In contrast, the evaluation set is fixed to 500 samples and does not vary with the pre-train data size. Finally, the pre-train corpus is constructed by combining training content and evaluation content.

\paragraph{SFT corpus construction.}

After constructing the synthetic KGs and the pre-training corpus (Figure~\ref{fig:Overall_framework}), we further build a supervised fine-tuning (SFT) corpus to endow the model with question-answering capability. Specifically, we generate QA pairs only from the train content of the pre-training corpus. For each sentence derived from a triple, we create a corresponding question and its answer; for example, from ``Name\#1 is the father of Name\#2'' we derive ``Q: Who is the father of Name\#2? A: Name\#1.'' Notably, we do not construct any SFT QA data from the evaluation content.

\subsection{Training setup}

Inspired by \cite{allen2023physics, zhang2025interplay}, we train GPT2-style AR LMs \cite{radford2019language} from scratch. Pre-training is conducted on the combined corpus of train and evaluation content. We control four experimental variables: $K=\texttt{num\_template}$, $N=\texttt{num\_train\_samples}$, $E=\texttt{num\_training\_epochs}$, and $L=\texttt{num\_layers}$. 

%Unless otherwise specified, experiments use the GPT2 configuration ($L$ layers, 12 heads, hidden size 768).

After pre-training, we perform SFT \cite{ouyang2022training} on the SFT corpus (generated from train content only) to equip the model with question-answering capability. No experimental variables are introduced during SFT. More details about the training setup are provided in Appendix~\ref{app:training_setup}.

\subsection{Evaluation setup}

As shown in Figure~\ref{fig:Overall_framework}, we evaluate models with three complementary query sets: Memorize QA and Logic QA are derived from the Evaluation Graph, while in-context learning Logic QA is derived from the ICL Graph. Example queries are provided in Table~\ref{tab:eval_icl_qa_examples}.

\paragraph{Memorize QA.}

Memorize QA queries facts explicitly present in Evaluation content, i.e., $\text{Q}_{\text{Mem}}$, to measure both memorization of pre-training text and basic QA capability after SFT.

\paragraph{Logic QA.}

Logic QA $\text{Q}_{\text{Logic}}$ evaluates whether the model can infer relational logic beyond memorization by querying facts that are absent from Evaluation content but logically implied by it. For instance, for an inversion pair (e.g., \texttt{father}/\texttt{son}), evaluation content may include only one direction such as ``A is the father of B'' but omit its inverse ``B is the son of A''; we then ask an evaluation question ``who is the son of A?'', whose correct answer is 
B. Success on such queries indicates that the model has internalized the inverse relation between \texttt{father} and \texttt{son} and can derive the missing counterpart from the observed direction. 

%Analogously, for a symmetric relation, $D_{\text{test}}$ contains only one direction and the query requires inferring the reversed statement from symmetry. 

\paragraph{In-context learning logic QA.}

In-context learning (ICL) logic QA $\text{Q}_{\text{ICL}}$ further tests whether the learned logic generalizes to unseen entities that never appear in Pre-train corpus.
Each prompt provides a context instantiating a relation and requires the model to produce its logically implied counterpart. We consider both completion-based evaluation $\text{Q}_{\text{ICL-C}}$, performed on the pre-trained model before SFT (e.g., ``A is the father of B. B is the son of \underline{\hspace{1em}}''), and QA-based evaluation $\text{Q}_{\text{ICL-Q}}$, performed after SFT (e.g., ``A is the father of B. Who is the son of A?''). This separation allows us to probe relational reasoning both in pure language modeling and in QA mode, while minimizing confounding effects from memorized entity associations.

\section{Result 1: Can auto-regressive LMs memorize and perform basic QA?}\label{Result1}

We use Memorize QA $\text{Q}_{\text{Mem}}$ as a prerequisite check to verify that models can memorize and answer factual queries, since failures on such queries would confound subsequent logic evaluations.

\begin{figure}[t]
\centering
\includegraphics[width=1\columnwidth]{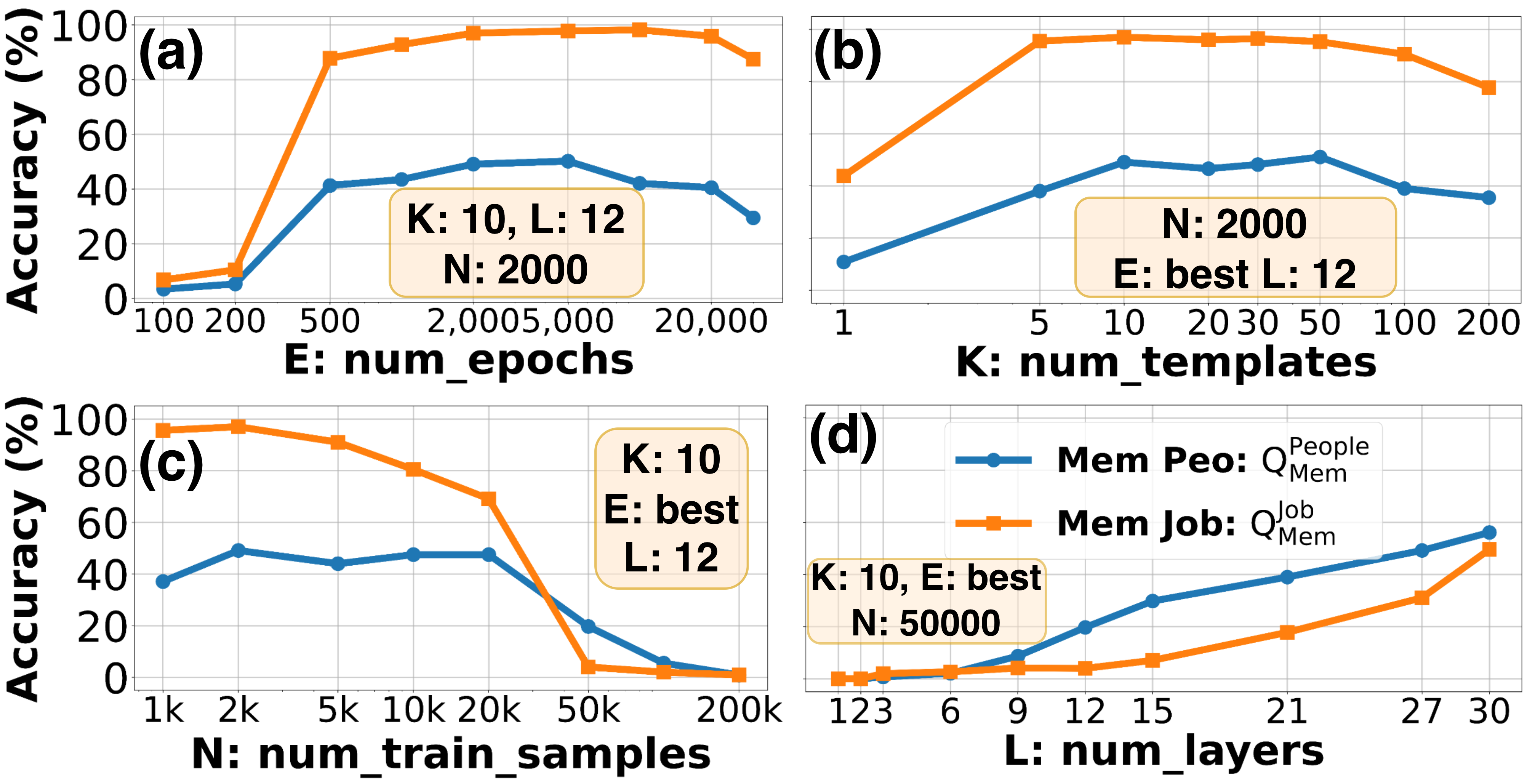}
\caption{Memorize QA ($\text{Q}_{\text{Mem}}$) evaluation under varying training epochs $E$, template size $K$, training sample size $N$, and model depth $L$. (a) Vary $E$ with $K,L,N$ fixed. (b) Vary $K$ with $N,L$ fixed. (c) Vary $N$ with $K,L$ fixed. (d) Vary $L$ with $K,N$ fixed.}
\label{fig:non_logic_figure}
\end{figure}

\paragraph{Task setting.}
We study four factors: $L$, $N$, $K$, and $E$. Unless otherwise stated, for each configuration we select the epoch $E$ that yields the highest accuracy (since the optimal epoch differs across $K$). Specifically, we (1) fix $L,N,K$ and sweep $E$; (2) fix $L,N$ and vary $K$; (3) fix $L,K$ and vary $N$; and (4) fix $K,N$ and vary $L$. We evaluate two Memorize QA subsets, QA-people $\text{Q}_{\text{Mem}}^{\text{People}}$ and QA-job $\text{Q}_{\text{Mem}}^{\text{Job}}$, whose answers are explicitly stated in evaluation content.

\paragraph{Observations.}

Figure~\ref{fig:non_logic_figure} summarizes the results. (a) With $K=10$ and $N=2000$, $\text{Q}_{\text{Mem}}^{\text{People}}$ and $\text{Q}_{\text{Mem}}^{\text{Job}}$ exhibit a rise-then-fall trend as $E$ increases; $\text{Q}_{\text{Mem}}^{\text{Job}}$ peaks above 95\% while $\text{Q}_{\text{Mem}}^{\text{People}}$ peaks around 50\%. (b) With $N=2000$, varying $K$ yields the same rise-then-fall pattern. (c) With $K=10$, increasing $N$ causes both accuracies to decrease monotonically, approaching zero for very large $N$ (e.g., $>10^5$). (d) Under $K=10$ and $N=50{,}000$, increasing depth $L$ consistently improves both accuracies, with gains persisting beyond 30 layers.

\paragraph{Takeaway and discussion.}
First, despite relatively low $\text{Q}_{\text{Mem}}^{\text{People}}$ accuracy (about 50\% at its best), $\text{Q}_{\text{Mem}}^{\text{Job}}$ reaches above 95\% under appropriate template diversity and training epochs, indicating that our pre-training plus SFT pipeline can support memorization and basic QA on the synthetic corpus, consistent with \citet{allen2023physics}. Second, when scaling the train sample size $N$ substantially, increasing model capacity (via larger $L$) becomes important to maintain and recover memorization and QA performance.

\section{Result 2: Can autoregressive LMs learn relational word logical semantics, and when?}

\begin{figure}[t]
\centering
\includegraphics[width=1\columnwidth]{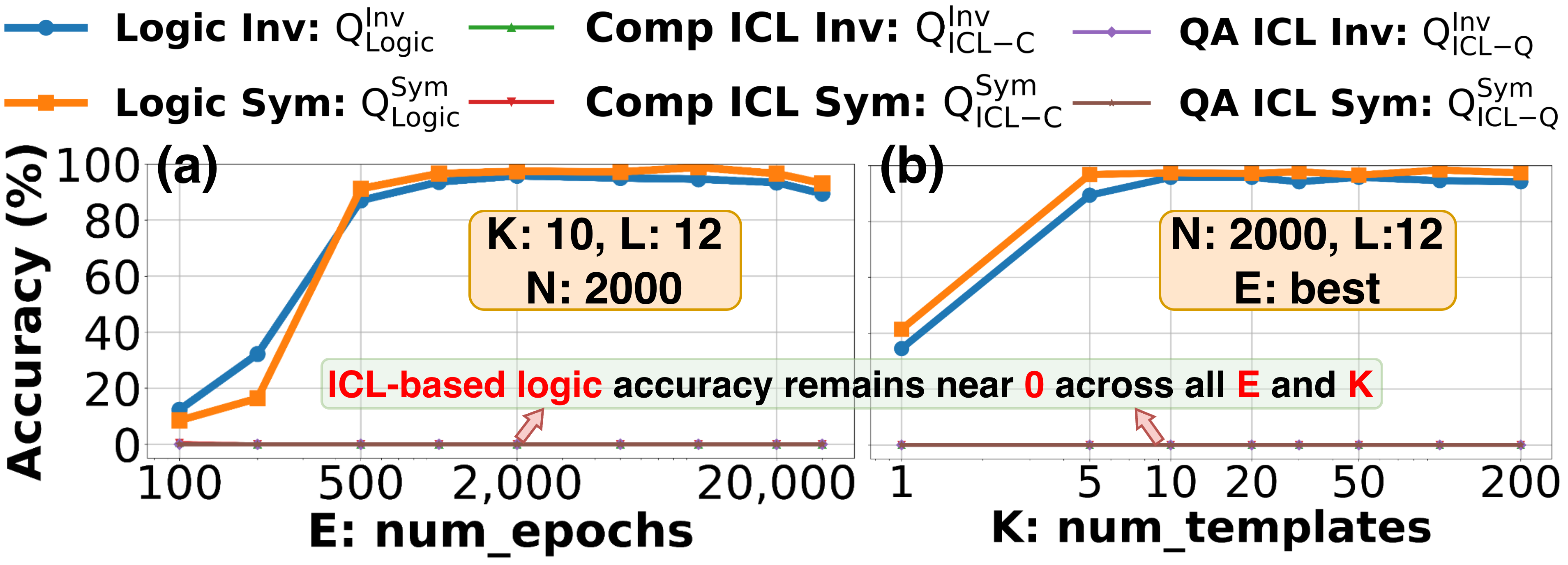}
\caption{Logic and ICL evaluation. (a) Vary $E$ with $K,L,N$ fixed. (b) Vary $K$ with $N,L$ fixed.}
\label{fig:logic_result_template_epochs}
\end{figure}

Having verified basic memorization and QA capability in Result 1, we now test whether GPT2-style auto-regressive LMs can learn the relational-word logical semantics inspired by KGs, such as inversion (e.g., \texttt{father/son}) and symmetry (e.g., \texttt{friend}). 
%We ask three questions: (1) can the model infer logically implied but unseen relation facts; (2) under what training conditions (data scale, training epochs, and model capacity) does this ability emerge; and (3) how do models that succeed differ internally from those that fail.
%We evaluate logical competence using Logic QA ($\text{Q}_{\text{Logic}}$) on held-out relation directions in evaluation content, and in-context learning logic QA on unseen entities via completion-based prompts ($\text{Q}_{\text{ICL-C}}$) (before SFT) and QA-based prompts ($\text{Q}_{\text{ICL-Q}}$) (after QA-based SFT).

\subsection{Task setting}
%We study whether relational logic emerges as a function of $N,K$ and $L$. 
%Unless otherwise specified, we evaluate logic using $\text{Q}_{\text{Logic}}$, $\text{Q}_{\text{ICL-C}}$, and $\text{Q}_{\text{ICL-Q}}$.
We ask three questions: (1) can the model learn the relational-word logical semantics; (2) under what training conditions ($N,K,L,E$) does this ability emerge; and (3) how do models that succeed differ internally from those that fail.

\paragraph{Task 1 (main sweeps).}
We use the same task setting as in Result~1 (the same sweeps over $E$, $K$, $N$, and $L$), but replace the evaluation with $\text{Q}_{\text{Logic}}$, $\text{Q}_{\text{ICL-C}}$, and $\text{Q}_{\text{ICL-Q}}$.

\paragraph{Task 2 (small-$L$ study).}
To isolate the effect of model depth, we additionally evaluate shallow models with $L\in\{1,2,3\}$. Unless stated otherwise, we fix $K=10$ and sweep $N$, reporting both $\text{Q}_{\text{Logic}}$ and $\text{Q}_{\text{ICL-C}}$.

\paragraph{Task 3 (layer-wise analysis).} \label{sec:layer_wise_analysis}
For prompts in $\text{Q}_{\text{ICL-C}}$ (e.g., ``A is the father of B. B is the son of \underline{\hspace{1em}}''), we extract, at each layer $l$, the next-token logit, softmax probability, and rank of the correct first token. We then compute layer-wise means over 1{,}800 prompts. (the full setup is provided in appendix~\ref{sec:detail_layer_wise_analysis})

\subsection{Observations.}

\paragraph{Task 1.}
Figure~\ref{fig:logic_result_template_epochs} (a, b) shows that for fixed $L$ and $N$, varying training epochs $E$ or template count $K$ yields a consistent pattern: $\text{Q}_{\text{Logic}}^{\text{Inv}}$ and $\text{Q}_{\text{Logic}}^{\text{Sym}}$ rise with training and reach stable peak accuracy above 95\%, while both $\text{Q}_{\text{ICL-C}}$ and $\text{Q}_{\text{ICL-Q}}$ remain near 0\%. 

In contrast, when sweeping data scale $N$ with fixed $L$ and $K$ (Figure~\ref{fig:emergence_reverse} (a)), $\text{Q}_{\text{Logic}}^{\text{Inv}}$ and $\text{Q}_{\text{Logic}}^{\text{Sym}}$ decrease as $N$ increases and drop below 5\% at $N=200{,}000$. Meanwhile, in-context performance improves nonlinearly: $\text{Q}_{\text{ICL-C}}$ exhibits a clear \textbf{phase transition} when $N$ exceeds 2{,}000, and $\text{Q}_{\text{ICL-Q}}$ shows a phase transition once $N$ exceeds 20{,}000.
%at $N=200{,}000$, it continues to increase without an obvious plateau.

\begin{figure*}[t]
\centering
\includegraphics[width=1\columnwidth]{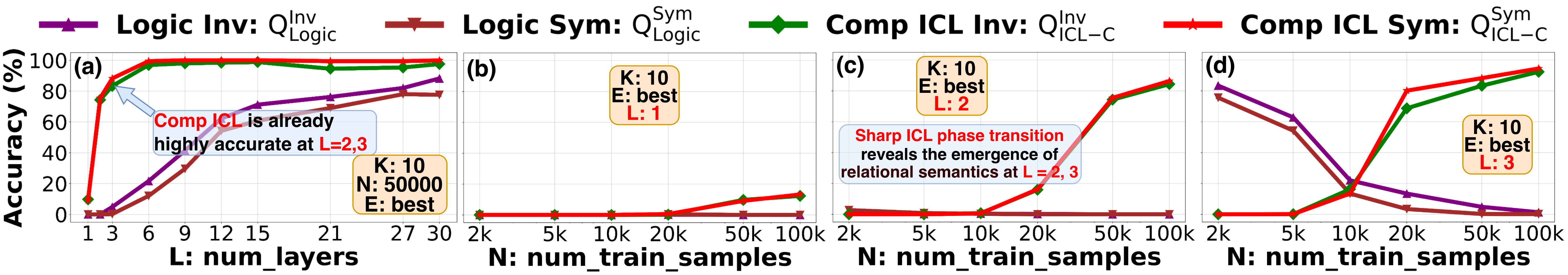}
\caption{Effect of model depth on logic and ICL completion evaluations. (a) In-context performance as $L$ varies under fixed $K$ and $N$. (b--d) For shallow models ($L=1,2,3$), evaluation performance across $N$ with fixed $K$.}

\label{fig:logic_layers}
\end{figure*}

Figure~\ref{fig:logic_layers} (a) further shows that increasing model depth $L$ (with fixed $K$ and $N$) improves in-context performance: $\text{Q}_{\text{ICL-Q}}$ (both inversion and symmetry) increases with $L$ and grows more slowly beyond $L\approx 12$. Notably, $\text{Q}_{\text{ICL-C}}$ improves sharply even for shallow models, reaching about 85\% accuracy at $L=3$ and around 10\% at $L=1$.

\begin{figure*}[t]
\centering
\includegraphics[width=1\columnwidth]{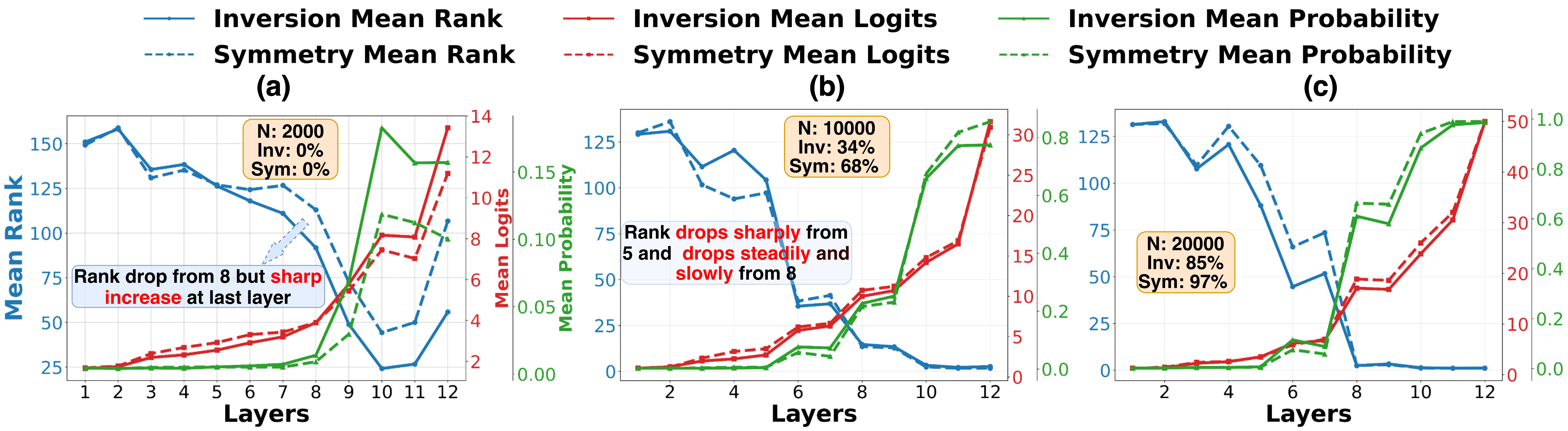}
%\caption{Layer-wise logit, probability, and rank of the first correct token for $\text{Q}_{\text{ICL-C}}$, averaged over 1{,}800 prompts, for models trained with different $N$.}
\caption{Layer-wise mean logit, mean probability, and mean rank of the first correct token on $\text{Q}_{\text{ICL-C}}$, for models trained with different $N$ (with $K=10$ and $L=12$ fixed). Panels (a–c) correspond to $N=2{,}000$, $10{,}000$, and $20{,}000$, respectively; the $\text{Q}_{\text{ICL-C}}$ accuracy for each setting is annotated in the figure.
}
\label{fig:logits}
\end{figure*}

\paragraph{Task 2.}
Figure~\ref{fig:logic_layers} (b, c, d) shows distinct behaviors for shallow models. With $L=1$, both $\text{Q}_{\text{Logic}}$ and $\text{Q}_{\text{ICL-C}}$ are low. 
%$\text{Q}_{\text{ICL-C}}$ reaches only about 15\% even when $N>50{,}000$. 
With $L=2$, $\text{Q}_{\text{Logic}}$ stays near 0\% across $N$, but $\text{Q}_{\text{ICL-C}}$ exhibits a \textbf{phase transition} once $N>10{,}000$ and can rise to about 85\%. With $L=3$, $\text{Q}_{\text{Logic}}$ can be high at $N=2{,}000$ but decreases as $N$ grows, and $\text{Q}_{\text{ICL-C}}$ follows a similar trend to the $L=2$ case.

\paragraph{Task 3.}
Figure~\ref{fig:logits} shows layer-wise analysis for ICL completion $\text{Q}_{\text{ICL-C}}$. When $N=2{,}000$ (Figure~\ref{fig:logits} (a)),
the mean rank stays above $\sim125$ until layer 8, improves at layers 8--10, and then drops sharply at layers 11--12. Mean probability mirrors this trend, rising only to about 15\% before dropping in the final layers. Mean logits increase slowly before layer 8 and then rise rapidly afterward.

In contrast, when $N=10{,}000$ and $20{,}000$ (Figure~\ref{fig:logits} (b, c)), mean rank improves sharply starting around layer 5 and continues to improve, reaching near rank $=1$ by layer 12; mean probability increases accordingly and approaches 100\% at the final layer. 
%Mean logits increase throughout and end substantially higher than in the $N=2{,}000$ setting.

\subsection{Takeaway.}
\paragraph{AR LMs can learn the logical semantics of relational word.}
Under appropriate training conditions, $\text{Q}_{\text{Logic}}$ and $\text{Q}_{\text{ICL-C}}$ exceed $95\%$, and $\text{Q}_{\text{ICL-Q}}$ reaches $\sim$85\% and keeps improving with larger $N$ without a clear plateau, as shown in Figure~\ref{fig:emergence_reverse}(a).

\paragraph{Relational semantics emerge with sufficient logic-bearing supervision, enabling generalization to unseen entities.}

This emergence exhibits a sharp phase transition from near-zero to near-perfect performance as the training data increases (Figure~\ref{fig:emergence_reverse}(a)), and it can arise even in small, shallow models with only 2--3 layers (Figure~\ref{fig:logic_layers}(c,d)).

%First, increasing the number of training samples that instantiate the target relations is necessary: performance remains near zero at small $N$ and exhibits a clear phase transition once $N$ crosses a threshold (e.g., around $N\approx 2{,}000$ in our setup). Second, model depth matters: while a small number of layers can already support in-context completion for relational logic, robust memorization and QA behavior require deeper models (e.g., at least three layers), and larger $N$ further increases the required capacity for strong generalization.

\paragraph{Successful relational generalization is associated with stable, logic-relevant representations in intermediate layers, whereas unsuccessful models exhibit late-stage, unstable signals that degrade at the final layer.} \label{takeaway:layer-wise}

From Figure~\ref{fig:logits}, we observe that high $\text{Q}_{\text{ICL-C}}$ accuracy coincides with an early improvement of the first-correct-token signal around layers 5--8, whereas low-performing models show a delayed onset (around layers 9--11) followed by a sharp final-layer degradation, as indicated by the layer-wise rank and probability curves.

% Models that achieve high in-context accuracy typically form strong signals in intermediate layers (roughly layers 5--8) and remain stable through later layers. In contrast, low-performing models show delayed and unstable behavior that only appears around layers 9--11 and can sharply deteriorate at the last layer, consistent with the abrupt changes observed in the layer-wise rank/probability curves.

\subsection{Discussion.}
\paragraph{Emergence of relational semantics and generalization}
We observe a phase transition phenomenon in $\text{Q}_{\mathrm{ICL}}$ once the number of inversion/symmetry training samples exceeds a threshold. This is consistent with prior reports of emergent capabilities with scale \cite{wei2022emergent}, but we suggest that the effective trigger can be the scale of targeted logical evidence rather than model size alone. A plausible explanation is a strategy shift: below the threshold the model relies on local co-occurrence or weak memorization, whereas above it the logic signal is frequent and diverse enough to support a reusable rule mechanism, in line with grokking-style dynamics \cite{nanda2023progress}. The opposite trends of $\text{Q}_{\mathrm{Logic}}$ (decreasing) and $\text{Q}_{\mathrm{ICL}}$ (increasing) with larger $N$ are consistent with $\text{Q}_{\mathrm{Logic}}$ being more affected by entity-memorization interference, while $\text{Q}_{\mathrm{ICL}}$ more directly reflects relational-rule induction.

\paragraph{Layer-wise correlates of relational generalization}
The Takeaway~\ref{takeaway:layer-wise} from our layer-wise analysis matches the residual-stream/circuits view in which intermediate layers are the primary locus where features are written and composed into task mechanisms, while later layers mainly read out these mechanisms and align them to the output distribution \cite{elhage2021mathematical}. This interpretation is also related to logit lens and tuned lens analyses, which decode intermediate hidden states into output-space predictions to trace how model beliefs evolve across layers \cite{nostalgebraist2020logitlens,belrose2023eliciting}. Accordingly, when relational logic is truly learned, the corresponding mechanism can be formed early and then amplified consistently; when it is weak or not fully integrated, late layers may rely on unstable heuristics and the final readout can suppress the partial logic signal due to competition with other features or distributional alignment. Combined with mechanistic grokking accounts \cite{nanda2023progress}, this suggests that failing models remain in a memorization/weak-mechanism regime, producing the observed ``improve-then-drop'' behavior near the output.

\paragraph{Grokking of relational semantics with $E$.}

\begin{figure}[t]
\centering
\includegraphics[width=1\columnwidth]{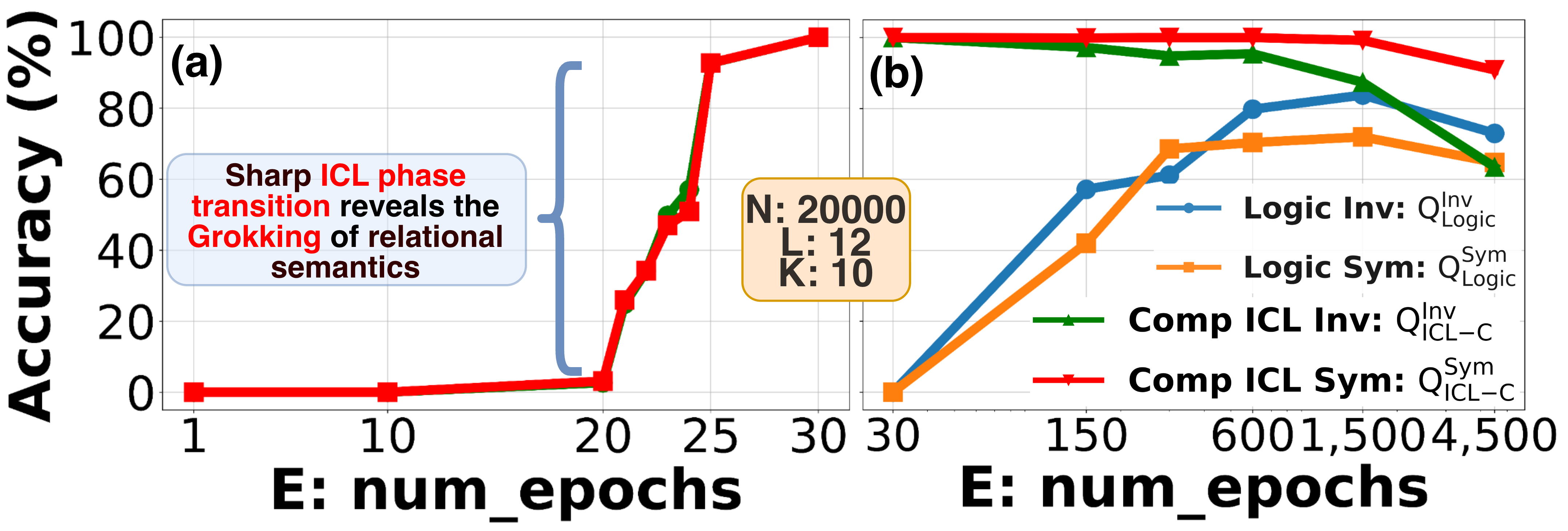}
\caption{Learning dynamics of $\text{Q}_{\text{Logic}}$ and $\text{Q}_{\text{ICL-C}}$ across training steps $E$ under fixed $N,L,K$, with panels showing early and later 30 training epochs.}
\label{fig:grokking}
\end{figure}

Figure~\ref{fig:grokking} shows how $Q_{\text{Logic}}$ and $Q_{\text{ICL}}$ change with training steps $E$ under fixed $N=20000$, $L=12$, and $K=10$. Panel (a) reveals a clear grokking transition: $\text{Q}_{\text{ICL-C}}^{\text{Inv}}$ and $\text{Q}_{\text{ICL-C}}^{\text{Sym}}$ abruptly jump to near-perfect accuracy around $E \approx 20$. 
Notably, this ICL Logic QA generalization, which is largely decoupled from direct memorization, emerges earlier than the improvement in memorization-dependent Logic QA. This ordering contrasts with the standard characterization of grokking as delayed generalization after memorization \cite{wang2024grokked,power2022grokking,wu2025rote}, and instead suggests that for relational semantics in our setting, generalization can precede memorization.

%This indicates that with sufficiently large $N$, continued training can trigger grokking of logical relational semantics.

%Our results align with prior grokking work and suggest a classic “circuit switch”: training shifts the model from memorization/shortcuts to a relational rule circuit, producing a sharp in-context generalization jump at a specific step\cite{wang2024grokked,nanda2023progress,power2022grokking}. Moreover, we provide the first evidence in a logic-focused setting (rather than a standard QA task) that grokking emerges only when the logic training data size exceeds a threshold.

\paragraph{Effect of template diversity on generalization and memorization.}

In the main experiments, we study the effect of $K$ only under the small-data setting ($N=2{,}000$). To test whether template diversity still matters when $N$ is already large enough to support strong generalization, we additionally fix $N=10{,}000$ and vary $K \in \{1,3,5,10\}$. As shown in Table~\ref{tab:template_diversity_10000}, both $Q_{\text{ICL}}^{\text{Inv}}$ and $Q_{\text{ICL}}^{\text{Sym}}$ are low when $K$ is small, improve substantially at $K=5$, and exceed $95\%$ at $K=10$. This shows that strong relational generalization requires not only enough training samples, but also sufficient template diversity. Moreover, Figure~\ref{fig:logic_result_template_epochs} (b) shows that $Q_{\text{Logic}}$ is also weak when $K$ is small, suggesting that template diversity affects memorization as well as generalization.

\begin{table}[t]
\resizebox{1\textwidth}{!}{\renewcommand{\arraystretch}{1}
\centering
\small
\begin{tabular}{c|cccc}
\hline
Template $K$ & 1 & 3 & 5 & 10 \\
\hline
$Q_{\text{ICL-C}}^{\text{Inv}}$ (\%) & 13.50 & 30.22 & 66.61 & 95.00 \\
$Q_{\text{ICL-C}}^{\text{Sym}}$ (\%) & 18.67 & 52.00 & 89.56 & 97.00 \\
\hline
\end{tabular}}
\caption{Effect of template diversity $K$ on in-context generalization when the number of training samples is fixed at $N=10{,}000$.}
\label{tab:template_diversity_10000}
\end{table}

\paragraph{Why does $\text{Q}_{\text{Logic}}$ decrease at large $N$?}
As shown in Figure~\ref{fig:emergence_reverse}, $\text{Q}_{\text{Logic}}$ decreases at large $N$. A likely reason is that larger $N$ requires memorizing more entity-level facts, especially many distinct names, while model size is fixed. Figure~\ref{fig:logic_layers}(a) further shows that increasing the number of layers substantially improves $\text{Q}_{\text{Logic}}$ at large $N$, suggesting that the drop mainly comes from limited model capacity rather than failed relational rule learning.

\begin{figure}[t]
\centering
\includegraphics[width=1\columnwidth]{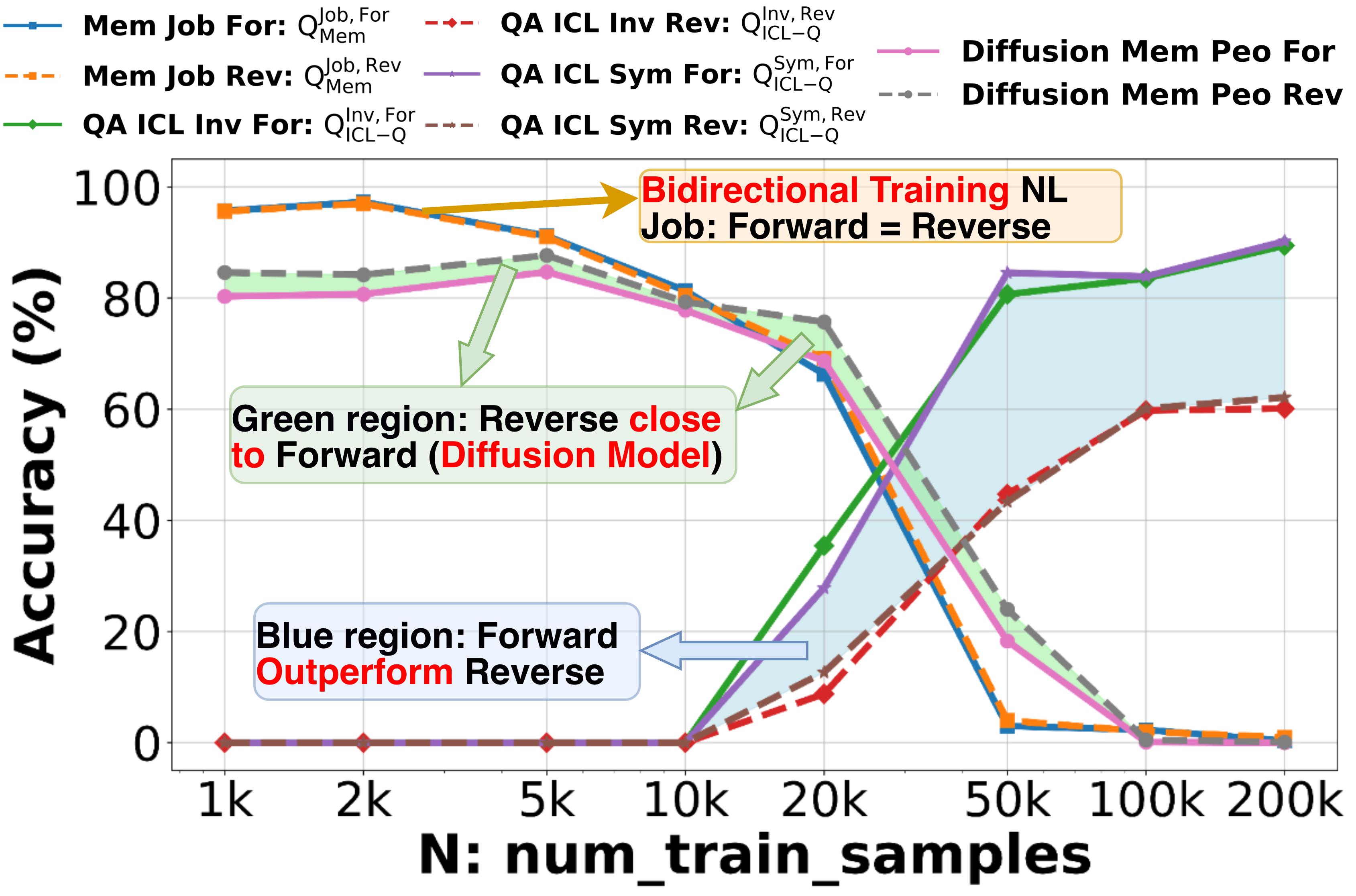}
\caption{Forward--reverse accuracy comparisons for  $\text{Q}_{\text{Mem}}^{\text{Job}}$ and  $\text{Q}_{\text{ICL-Q}}$ , together with Memorize-people QA results under a diffusion model. All experiments vary $N$ with $K$ and $L$ fixed.}
\label{fig:reverse_other}
\end{figure}

%\section{Result 3: When Does the Reversal Curse Occur? Word Order vs.\ Missing Relational Semantics}
\section{Result 3: What drives reversal-type failures: relational semantics or order bias?}\label{sec:result-reversal}
Prior work uncovered reversal curse \cite{berglund2023reversal}.
However, it remains an open question whether this failure is primarily caused by order bias in AR decoding, or by the model's inability to internalize the inversion semantics between relational words (e.g., \texttt{father} vs.\ \texttt{son}).
Most existing evaluations support the former explanation, but do not fully rule out the latter.
Building on our Results 1--2, we design contrastive experiments that isolate order bias while controlling inversion relational semantics.

\subsection{Task setting}

\paragraph{Forward vs.\ Reverse queries.}
Given a fact ``A is the father of B'' (order: $\text{A}\!\rightarrow\!\texttt{father}\!\rightarrow\!\text{B}$), a \textbf{Reverse} query conditions on B and asks for A (``Who is the father of B?'' $\Rightarrow$ A), whereas a \textbf{Forward} query conditions on $A$ and asks for $B$ (``A is the father of who?'' $\Rightarrow$ B). 

\textbf{Nuance ($\text{Q}_{\text{Logic}}$).}
%If evaluation content provides only ``A is the father of B'' ($A\!\rightarrow\!\texttt{father}\!\rightarrow\!B$), the inversion query $\text{Q}_{\text{logic}}^{\text{Inv}}}$ ``Who is the son of A?'' $\Rightarrow$ B matching the training order ($A\!\rightarrow\!\cdot\!\rightarrow\!B$); thus it is a \textbf{Forward} query w.r.t.\ information order. Symmetry query$\text{Q}_{\text{logic}}^{\text{Sym}}}$.
If the evaluation content provides only ``A is the father of B'' ($A\!\rightarrow\!\texttt{father}\!\rightarrow\!B$), the inversion logic query $\text{Q}_{\text{Logic}}^{\text{Inv}}$ (``Who is the son of A?'' $\Rightarrow$ B) matches the training order ($\text{A}\!\rightarrow\!\cdot\!\rightarrow\!\text{B}$). 
Therefore, it is a \textbf{Forward} query with respect to information order, rather than a Reverse query like $\text{Q}_{\text{Mem}}^{\text{People}}$.
The same logic applies to the symmetry query $\text{Q}_{\text{Logic}}^{\text{Sym}}$.

\paragraph{Task 1: One-directional data exposure for relational attributes (people). }
All person--person evaluation facts use the one-directional template ``A is the \{relation\} of B'' ($A\!\rightarrow\!r\!\rightarrow\!B$). We evaluate (i) Memorize-people QA $\text{Q}_{\text{Mem}}^{\text{People}}$, (ii) inversion logic via $\text{Q}_{\text{Logic}}^{\text{Inv}}$ and $\text{Q}_{\text{ICL-Q}}^{\text{Inv}}$, and (iii) symmetry logic via $\text{Q}_{\text{Logic}}^{\text{Sym}}$ and $\text{Q}_{\text{ICL-Q}}^{\text{Sym}}$. All evaluations are conducted using both Forward and Reverse query formats.

\paragraph{Task 2: Bidirectional data exposure for non-relational attributes (job).}
For job facts, we include both forward and reverse surface templates in training (e.g., ``chef is the job of A'' and ``A's job is chef'') and evaluate $\text{Q}_{\text{Mem}}^{\text{Job}}$ using matched Forward/Reverse queries. Details are shown in Table~\ref{tab:templates}.

\paragraph{Task 3: Comparison with large language diffusion model.}
%To investigate whether the ''reversal curse'' stems from a lack of logical representation or the AR (AR) generation order, we introduce a Diffusion Language Model baseline and compare its learning dynamics against the AR model. We employ a Bidirectional Transformer backbone trained with the LLaDA framework \cite{nie2025large} (Bernoulli Masking for pre-training, Answer-Only Masking for SFT). Aside from the architecture and training objective, other configurations remain consistent with the AR baseline.

%To test whether the reversal curse arises from missing logical representations or AR order bias, 
We train a diffusion LM baseline on our synthetic data using a bidirectional Transformer under LLaDA \cite{nie2025large} and compare its dynamics to an AR model (see Appendix~\ref{app:diffusion_model} for settings).

\subsection{Observations.}
\paragraph{Tasks 1.}
Across both Memorize-people and Logic (inversion and symmetry) queries, the Reverse accuracy curves are far below their Forward counterparts (Figure~\ref{fig:emergence_reverse} (b) and Figure~\ref{fig:reverse_other}).
Moreover, within the same query format (forward or reverse), Symmetry and Inversion consistently perform at similar levels (e.g., $\text{Q}_{\text{Logic}}^{\text{Sym,For}}$ vs.\ $\text{Q}_{\text{Logic}}^{\text{Inv,For}}$, or $\text{Q}_{\text{Logic}}^{\text{Sym,Rev}}$ vs.\ $\text{Q}_{\text{Logic}}^{\text{Inv,Rev}}$).

\paragraph{Task 2.}
The forward and reverse accuracies on $\text{Q}_{\text{Mem}}^{\text{Job}}$ become nearly identical (Figure~\ref{fig:reverse_other}, $\text{Q}_{\text{Mem}}^{\text{Job,For}}$ vs.\ $\text{Q}_{\text{Mem}}^{\text{Job,Rev}}$).
%This contrast indicates that reversal-type failures largely stem from \emph{order mismatch} between the training surface form and the query conditioning pattern, rather than from an inability to encode the underlying relational semantics.

\paragraph{Task 3.}

Figure~\ref{fig:reverse_other} illustrates that, in the diffusion model, forward- and reverse-based Memorize-people query accuracies are nearly identical; intriguingly, the reverse setting is on average about 4\% higher than the forward setting.

%Figure~\ref{fig:reverse_other} illustrates the training dynamics. Throughout the training process, there is no observable performance gap between the two directions, with both converging to a high accuracy range ($\sim$90\%) simultaneously. This contrasts sharply with the distinct gap observed in AR models.

\subsection{Takeaway}

\paragraph{Reversal-type failures primarily stem from order bias in left-to-right AR decoding rather than missing relational semantics.}

Symmetry and inversion perform similarly under both formats, indicating comparable relational semantics. Since symmetry should not induce reversal failures, the gap is best explained by order sensitivity rather than deficient inversion semantics.

%Symmetry and inversion exhibit nearly identical performance distributions under both forward and reverse formats, suggesting comparable relational semantics. Because symmetry semantics is not expected to cause reversal-type failures, this parity implies that such failures are better explained by order sensitivity than by deficient inversion semantics.

%Models perform well when the query conditions on information in the same surface order as seen during training, but fail when the conditioning order is reversed.
%This is supported by: (i) Result~2, which shows the model can learn inversion semantics under sufficient training; (ii) Tasks~1, where inversion and symmetry exhibit nearly identical Forward--Reverse gaps (high Forward, low Reverse), indicating a rule-agnostic cause; and (iii) $\text{Q}_{\text{non}}^{\text{people}}$, where Reverse queries remain much worse than Forward despite querying facts explicitly present in evaluation content.

%\paragraph{bidirectional data exposure substantially mitigates reversal failures.}
%With bidirectional data exposure in Task~2 and diffusion model, Forward and Reverse accuracies on are nearly identical.
%showing that exposing both orders during training effectively reduces the reversal curse.
\paragraph{Bidirectional data exposure substantially mitigates reversal failures.}
When the same autoregressive model is trained with both forward and reverse surface forms, the Forward and Reverse accuracies become nearly identical, indicating that the forward--reverse gap is largely removed.

\subsection{Discussion}
\paragraph{Reversal failures reflect AR order bias, not semantics}
Prior work reports a consistent reversal curse \cite{berglund2023reversal}. Our controlled KG-based experiments further support the view that this failure is primarily due to order bias \cite{lv2024analysis, zhu2024towards}. Crucially, unlike many studies on web-scale pre-trained models where semantic confounds are difficult to rule out, we disentangle order bias from inversion relational semantics and, for the first time, verify that the reversal curse is not caused by a failure to understand inversion semantics.

\paragraph{Bidirectional data exposure mitigates the reversal curse}
We also observe a direct mitigation consistent with the order-based account. 
In Task~2, bidirectional data exposure makes forward and reverse accuracies on $\text{Q}_{\text{non}}^{\text{job}}$ nearly identical, indicating that training exposure to both orders largely removes reversal failures. 
This complements prior mitigation efforts that adjust training signals \cite{lv2024analysis} and is consistent with recent evidence that non-AR diffusion LMs are less sensitive to order reversal, further implicating left-to-right decoding bias as the root cause \cite{nie2025large}.

\section{Result 4: How general is the observed emergence phenomenon beyond the KG-synthetic from-scratch setting?}

\subsection{Extension to pretrained models}

\paragraph{Task setting.}
To test whether our findings extend beyond from-scratch training, we conduct an additional experiment on pretrained GPT-2 small (12 layers $\times$ 12 heads). Starting from the pretrained checkpoint, we continue training the model using only the KG-synthetic pre-training corpus, without QA-style SFT, and evaluate inversion and symmetry under the same ICL completion setting as in the main experiments. We vary the number of training samples $N \in \{0, 10\mathrm{k}, 20\mathrm{k}, 30\mathrm{k}, 50\mathrm{k}\}$, where $N=0$ denotes the pretrained model without continue training.

\begin{table}[t]
\resizebox{1\textwidth}{!}{\renewcommand{\arraystretch}{1}
\centering
\begin{tabular}{c|ccccc}
\hline
$N$ & 0 & 10,000 & 20,000 & 30,000 & 50,000 \\
\hline
$\text{Q}_{\text{ICL-C}}^{\text{Inv}}$ & 6.00 & 20.23 & 35.34 & 37.53 & 98.89 \\
$\text{Q}_{\text{ICL-C}}^{\text{Sym}}$ & 8.00 & 23.37 & 37.28 & 38.34 & 99.32 \\
\hline
\end{tabular}}
\caption{Completion-based ICL accuracy $\text{Q}_{\text{ICL-C}}$ of pretrained GPT-2 small after continued training on the KG-synthetic corpus with different numbers of training samples $N$. %A sharp transition appears when $N$ reaches 30,000, mirroring the phase-transition-like behavior observed in from-scratch training.
}
\label{tab:pretrained_gpt2}
\end{table}

\paragraph{Observations.}
Without adaptation to the KG-synthetic corpus ($N=0$), pretrained GPT-2 performs near chance. Accuracy rises gradually with more $N$, remains below $40\%$ up to $N=30\mathrm{k}$, and then jumps to about $99\%$ at $N=50\mathrm{k}$.

\paragraph{Takeaway and discussion.}
The pretrained model shows the same phase-transition-like behavior as the from-scratch setting, suggesting that this pattern is not specific to random initialization on synthetic data. Instead, even pretrained autoregressive models require sufficient rule-consistent supervision to internalize and generalize relational semantics.

\subsection{Extension to more complex relations}

To test whether the emergence behavior in Result 2 extends beyond symmetry and inversion, we study a more complex compositional relation: $A$ is father of $B$, $B$ is father of $C$ $\Rightarrow$ $A$ is grandfather of $C$. Using the same KG-synthetic setup, we evaluate unseen-entity generalization with $Q_{\text{ICL-C}}$ without SFT. When $N=50{,}000$, accuracy remains near $0\%$; when $N=100{,}000$, it rises sharply to $47.6\%$. Although this is lower than symmetry/inversion (about $95\%$) and requires a larger $N$, the qualitative pattern is similar: the model first fails to generalize the rule, then improves sharply once sufficient rule-consistent supervision is provided. This suggests that the emergence phenomenon is not specific to symmetry or inversion, but can generalize to more complex logical relational structures, albeit with higher data requirements and lower final accuracy.

\section{Related work}
\paragraph{Controlled training from scratch with synthetic data}
Controlled, from-scratch training on synthetic data is increasingly used to attribute LM behaviors under fully known data-generating processes and to avoid confounds such as data contamination \cite{zhang2025interplay,liu2025prorl,yuan2025f}.
Such testbeds, exemplified by the Physics of LMs line and curated synthetic corpora like TinyStories, isolate specific capabilities and failure modes while making small-scale model development more interpretable \cite{allen2023physics, eldan2023tinystories}.
%Following this controlled-training philosophy, we propose a KG-grounded synthetic framework that generates text from symmetry/inversion triples and trains GPT-style LMs from scratch to study relational-logic emergence and generalization.

\paragraph{Ability emergence phenomenon.}
Prior work reports threshold-like “emergent” gains in reasoning with scale and shows that training can exhibit phase-transition dynamics in generalization \cite{wei2022emergent, nanda2023progress}.
Mechanistic analyses further suggest that transformers build computation by progressively composing features in the residual stream, with intermediate layers often forming stable rule-like circuits that explain abrupt behavioral shifts \cite{elhage2021mathematical}.

\paragraph{Reversal curse}
The reverse curse phenomenon was introduced by \citet{berglund2023reversal}. Subsequent analyses attribute it to directional training signals induced by causal masking and propose training-time interventions that expose alternative conditioning patterns \cite{lv2024analysis, zhu2024towards}. Lin et al.~\citep{lin2024delving} further show that reversal generalization is highly sensitive to the structure of training documents, and interpret this effect through an inherent bias in fact recalling. In parallel, LLaDA \cite{nie2025large} suggests that departing from the standard AR paradigm can help mitigate this effect.

%Through controlled training experiments, we show that the reverse curse is primarily driven by order sensitivity, rather than an inability of the model to learn the inversion logic of relational words.

\section{Conclusion}
We train AR LMs from scratch with KG-based synthetic data to address two questions: (i) Can AR LMs learn relational word logical semantics, and when? (ii) What drives reversal-type failures?
%: deficient relational semantics or order bias? 
In our results, we observe a sharp phase transition in which relational semantics emerge with sufficient logic-bearing supervision, even in shallow (2–3 layer) models.
%, and successful generalization aligns with stable intermediate-layer signals. 
Moreover, order-matched forward/reverse tests indicate that reversal failures are primarily driven by AR order bias rather than deficient inversion semantics.

\clearpage

%.............................Limitations

\section*{Limitations}
\paragraph{Limited relational word types}
Our study focuses on a narrow set of relation properties—primarily symmetric relations and inverse kinship pairs (plus a simple person–job attribute). While these cover two canonical forms of relational logic, they do not represent the broader landscape of relational semantics encountered in natural language and knowledge graphs. In particular, we do not test properties such as transitivity, antisymmetry, hierarchical relations, multi-hop compositional rules, or interactions among multiple properties, which may exhibit different emergence thresholds and failure modes.

\paragraph{Template-based language generation}

Although we use multiple verbalization templates and shuffle sentence order to increase surface diversity, the corpus is still template-generated and thus may not fully reflect the breadth of paraphrastic variation and contextual nuance in natural text. This limitation does not invalidate the controlled comparisons in our setting, but it may affect how directly the quantitative thresholds and emergence dynamics transfer to more naturalistic corpora.

\paragraph{Large language diffusion model training}

Training diffusion LMs from scratch on our synthetic corpus is not yet fully mature in our implementation, particularly for the supervised fine-tuning stage and its associated procedures. Therefore, our diffusion results currently support only the qualitative conclusion that forward and reverse queries exhibit little gap under diffusion-style training. A comprehensive accuracy-level comparison between diffusion and autoregressive models across the full set of query types remains future work.

\paragraph{ICL evaluation shortcut}

Although our ICL setting tests generalization to unseen entities, the prompts involve only a small number of entities and explicitly provide a forward relational statement in context. As a result, some questions may be solvable via shallow matching or local entity correspondence, without requiring a fully abstract understanding of relational semantics. Therefore, our ICL results should be interpreted as evidence of rule-like generalization beyond memorization, but not as a definitive test of pure relational-semantic understanding.

\paragraph{Limited pretrained-model evaluation}
Our extension to pretrained GPT-2 small remains limited in scope. Specifically, we evaluate only ICL completion accuracy, and do not examine forward–reverse gaps or reversal-type failures in the pretrained setting. Therefore, while these results suggest that phase-transition-like emergence is not specific to from-scratch training, they do not establish whether the order-bias findings in Section 5 generalize to pretrained autoregressive models.

\section*{Acknowledgements}
This work was partially supported by JST SPRING JPMJSP2110 (YZ), JSPS KAKENHI 22H05106, 23H03355, and JST CREST JPMJCR21N3 (HS). It was also supported by the “Development Acceleration Use” program of ABCI 3.0, provided by AIST and AIST Solutions. We sincerely thank all reviewers for their valuable comments, constructive suggestions, and strong support during the rebuttal stage.

\section*{Ethics Statement}
This study complies with the \href{https://www.aclweb.org/portal/content/acl-code-ethics}{ACL Ethics Policy}.

%.............................References
\bibliography{references}

\clearpage

%.............................Appendix
\appendix
\section{Appendix}

\subsection{Training setup} \label{app:training_setup}

\subsubsection{Model architecture}
We train a decoder-only, GPT-2–style model \cite{radford2019language} from scratch on our KG-based synthetic corpus, using the standard GPT-2 tokenizer. Except for experiments where we vary the number of layers ($L$), we use a fixed architecture with 12 layers, 12 attention heads, and a 768-dimensional hidden size. On 8$\times$A100 (40GB), training a model with ($N=2000$) and ($K=10$) (about 2 million tokens) to its best performance takes approximately 1.2 hours.

\subsubsection{Hyperparameters}
\paragraph{Pre-training.} We use a batch size of 491{,}520 tokens per iteration, a learning rate of ($6\times10^{-4}$) with weight decay 0.1, cosine decay to a minimum learning rate of ($6\times10^{-5}$), and 500 warmup iterations. All models are trained in bf16 precision.

\paragraph{SFT.} We fine-tune with standard supervised fine-tuning \cite{ouyang2022training}, using a learning rate of ($3\times10^{-5}$) and a batch size of 32{,}768 tokens per iteration.

\paragraph{Inference.} We generate outputs with temperature 0.8 and top-k = 100.

\subsubsection{Dataset statistics and training configuration}

Table~\ref{tab:best_config_by_N} summarizes the best-performing training configurations for each dataset size ($N$) under fixed ($L=12$) and ($K=10$), including the resulting token counts, iterations per epoch, and the approximate pre-training/SFT schedules (epochs or iterations) and wall-clock time on 8$\times$A100 (40GB). Notably, both the optimal pre-training length and the overall runtime vary substantially with the corpus size, and all reported epochs and times are approximate.

Table~\ref{tab:eval_question_counts} reports the number of evaluation questions for each task type when the evaluation set is fixed to 500 samples. Since forward and reverse formats are constructed to be balanced, we list only the reverse counts for clarity.

\subsection{Supplementary material}
Unless otherwise stated, we report results from a single run due to computational constraints (training from scratch and fine-tuning are expensive and time-consuming under our GPU budget).

Moreover, all data are synthetically generated and do not contain personally identifying information. We also avoid generating offensive content by construction and manually spot-check a sample of the generated corpus. 
Finally, all datasets in this work are synthetically generated for research and reproducibility purposes. The released artifacts (code/data/model checkpoints, if any) are intended for research use only.

Our experiments were facilitated by leveraging \href{https://pytorch.org}{PyTorch},  \href{https://huggingface.co}{Huggingface}, and \href{https://numpy.org}{Numpy} as essential tools. Furthermore, we use \href{https://chat.openai.com/#}{ChatGPT} in our paper writing and programming.

We will release the code under the MIT License. The generated synthetic datasets (and trained checkpoints, if released) will be distributed under the same license terms specified in the repository. We use standard open-source libraries (e.g., PyTorch under BSD-3-Clause, NumPy under BSD-3-Clause, and Transformers under Apache-2.0) and comply with their licenses.

\subsection{Data generation framework}
We briefly described this in section~\ref{methodology} (Methodology), but the coverage was not sufficiently detailed or technical. In this section, we provide a more formal description of the KG-based corpus construction, along with detailed statistics for each dataset. In Table~\ref{tab:text_pipeline_examples}, we illustrate the end-to-end text generation pipeline with a concrete example.

\subsubsection{KG and triple generation}

As illustrated in Figure~\ref{fig:Overall_framework}, each knowledge graph (KG) instance corresponds to a single semantic sample. We define a KG as a triple $\mathcal{G}=(\mathcal{E},\mathcal{R},\mathcal{T})$, where $\mathcal{E}$, $\mathcal{R}$, and $\mathcal{T}$ denote the entity set, relation set, and triple set, respectively. The entity set is given by $\mathcal{E}=\mathcal{E}^{N}\cup\mathcal{E}^{J}$, where $\mathcal{E}^{N}=\{e_{1}^{N},e_{2}^{N},e_{3}^{N}\}$ consists of three name entities, and $\mathcal{E}^{J}=\{e_{1}^{J},e_{2}^{J},e_{3}^{J}\}$ consists of their corresponding job entities. The relation set $\mathcal{R}$ comprises three components: (i) three inversion relation pairs $\mathcal{R}_{\mathrm{inv}}=\{(r_{1}^{I},r_{1}^{I,-1}), (r_{2}^{I},r_{2}^{I,-1}), (r_{3}^{I},r_{3}^{I,-1})\} = \{(\texttt{father},\!\texttt{son}),\!(\texttt{husband},\!\texttt{wife}),\!(\texttt{uncle},\!\texttt{niece})\}$;
(ii) three symmetric relations $\mathcal{R}_{\mathrm{sym}}=\{r_{1}^{S},r_{2}^{S},r_{3}^{S}\} = \{\texttt{friend},\texttt{brother},\texttt{spouse}\}$;
and (iii) a single attribute relation $r_J=\texttt{job}$.

To construct the triple set $\mathcal{T}$, we randomly select one inversion pair $(r_{k}^{I},r_{k}^{I,-1})\in\mathcal{R}_{\mathrm{inv}}$ and one symmetric relation $r_{l}^{S}\in\mathcal{R}_{\mathrm{sym}}$, where $k,l\in\{1,2,3\}$. Each resulting KG contains exactly seven triples: an inversion-consistent pair $(e_{1}^{N},r_{k}^{I},e_{2}^{N})$ and $(e_{2}^{N},r_{k}^{I,-1},e_{1}^{N})$; a symmetry-consistent pair $(e_{2}^{N},r_{l}^{S},e_{3}^{N})$ and $(e_{3}^{N},r_{l}^{S},e_{2}^{N})$; and three job attribute triples $(e_{i}^{N},r_J,e_{i}^{J})$ for $i\in\{1,2,3\}$.

% Requires: \usepackage{booktabs}

\begin{table*}[t]
\resizebox{1\textwidth}{!}{\renewcommand{\arraystretch}{1}
\centering
%\small

\setlength{\tabcolsep}{6pt}
\renewcommand{\arraystretch}{1}
\begin{tabular}{r c c c c c}
\toprule
\textbf{$N$} & \textbf{\#Tokens (Million)} & \textbf{Num\_iters/Epoch} & \textbf{Best PT Epochs} & \textbf{PT Time (h)} & \textbf{Best SFT Iters} \\
\midrule
  2{,}000   &  2.0  &   4.16  & 2000 &  1.2 & 2000 \\
  5{,}000   &  4.7  &   9.56  & 1100 &  1.2 & 2000 \\
 10{,}000   &  9.14 &  18.59  & 1100 &  2.5 & 2000 \\
 20{,}000   & 18.0  &  36.6   & 1400 &  5.0 & 2000 \\
 50{,}000   & 53.8  & 109.0   &  460 &  5.0 & 3000 \\
100{,}000   & 89.0  & 180.0   & 1000 & 20.0 & 3000 \\
\bottomrule
\end{tabular}}

\caption{Best training configurations under fixed $L=12$ and $K=10$ for different dataset sizes $N$. Token counts and training times are approximate estimates measured on 8$\times$A100 (40GB). "PT" denotes Pre-train in the table.}
\label{tab:best_config_by_N}

\end{table*}

% Requires: \usepackage{booktabs}

\begin{table}[t]
\resizebox{1\textwidth}{!}{\renewcommand{\arraystretch}{1}
\centering

\setlength{\tabcolsep}{8pt}
\renewcommand{\arraystretch}{1.15}
\begin{tabular}{l r}
\toprule
\textbf{Task / Question Type (Reverse)} & \textbf{Count} \\
\midrule
QA People (Memorize)         & 1000 \\
QA Job (Memorize)            & 1500 \\
QA Inversion (logic)          &  500 \\
QA Symmetry (logic)           &  500 \\
ICL Completion (Inversion)    & 1800 \\
ICL Completion (Symmetry)     &  900 \\
ICL QA (Inversion)            & 1800 \\
ICL QA (Symmetry)             &  900 \\
\bottomrule
\end{tabular}}
\caption{Counts of evaluation questions under a fixed evaluation set of 500 samples. Forward and reverse formats are balanced; we report reverse counts only.}
\label{tab:eval_question_counts}
\end{table}

\subsubsection{Training corpus construction}

In the previous subsection, we constructed a synthetic KG whose relations explicitly exhibit inversion and symmetry. We now define how the KG triples are transformed into the pre-training corpora:

\[
\begin{aligned}
D_{\alpha}
&= \mathcal{P}_{\alpha}(N,K;\mathcal{F}) \\
&{\scriptstyle
= \Big\{\!\big\{
\mathrm{Shuffle}\big(\{\,f^{(\tau^{(k)}_{i})}(T^{(n)}_{i})\,\}_{i\in\mathcal{I}_{\alpha}}\big)
\;\big\}_{k=1}^{K}\!\Big\}_{n=1}^{N}
}, \\
&\quad \text{where } \tau^{(k)}_{i}
\overset{\text{i.i.d.}}{\sim}\mathrm{Unif}(\{1,2,3,4\}), \\
&\quad \mathcal{I}_{\alpha}=
\begin{cases}
\{1,2,3,4,5,6,7\}, & \text{if $\alpha$ = train},\\
\{1,3,5,6,7\},     & \text{if $\alpha$ = evaluation}.
\end{cases}
\end{aligned}
\]

Here \(D_{\alpha}\) denotes the training corpus and is generated by \(\mathcal{P}_{\alpha}\) parameterized by \(N=\texttt{num\_train\_samples}\), \(K=\texttt{num\_template}\), and the fixed set of four sentence formats \(\mathcal{F}=\{f^{(1)},f^{(2)},f^{(3)},f^{(4)}\}\)(Detail examples are shown in Table~\ref{tab:templates}). The outer set \(\{\cdot\}_{n=1}^{N}\) indexes the \(N\) semantic samples, where the \(n\)-th sample is defined by a tuple of several KG triples \(T^{(n)}_{i}\). For each semantic sample, we generate \(K\) different paragraphs expressing same information, indexed by \(\{\cdot\}_{k=1}^{K}\), each corresponding to a paragraph-level template choice. Within the \(k\)-th paragraph, each sentence \(i\in\mathcal{I}_{\alpha}\) is produced by first drawing a format index \(\tau^{(k)}_{i}\) independently and uniformly from \(\{1,2,3,4\}\), then transforming triple \(T^{(n)}_{i}\) with the selected format via \(f^{(\tau^{(k)}_{i})}(T^{(n)}_{i})\). Finally, \(\mathrm{Shuffle}(\cdot)\) randomly permutes the resulting several sentences to form the paragraph text, yielding \(K\) shuffled paragraphs per semantic sample and \(N\) samples in total for \(D_{\alpha}\).

We create two pre-training corpora with different completeness. Train content $D_{\text{train}}$ uses $\mathcal{I}_{\text{train}}$ (7 sentences) and thus contains full inversion/symmetry information; its sample count equals $N=\texttt{num\_train\_samples}$. Evaluation content $D_{\text{eval}}$  uses $\mathcal{I}_{\text{eval}}$ (5 sentences) by removing one direction of each inversion/symmetry pair, serving for evaluation. Train and evaluation content are generated independently, and the number of evaluation samples is fixed to $500$. Finally, we merge the train content $D_{\text{train}}$  and the evaluation content $D_{\text{eval}}$  to form the pre-training corpus for the LLM.

\subsubsection{SFT corpus construction}

As illustrated in Figure~\ref{fig:Overall_framework} and examples shown in Table~\ref{tab:text_pipeline_examples}, after constructing the synthetic KG and the pre-training corpora $D_{\text{train}}$ and $D_{\text{eval}}$, we build a supervised fine-tuning (SFT) corpus to equip the model with question-answering capability. We generate QA pairs only from the train content:
\[
B_{\text{train}}
=\mathcal{S}(D_{\text{train}};\mathcal{U})
=\Big\{\;\{(\text{Q}_{n,i},A_{n,i})\}_{i=1}^{7}\;\Big\}_{n=1}^{N},
\]
where $N=\texttt{num\_train\_samples}$ and $\mathcal{U}=\{u^{(1)},u^{(2)},u^{(3)},u^{(4)}\}$ is a set of four question templates. Concretely, for each train semantic sample $n$, we construct seven QA pairs $\{(\text{Q}_{n,i},A_{n,i})\}_{i=1}^{7}$, each derived from one of its seven underlying triples, with answers deterministically extracted from the queried triple. Importantly, we do not generate any QA pairs from the evaluation content (i.e., $\mathcal{Q}_{\text{eval}}=\varnothing$); since evaluation queries are posed on $D_{\text{eval}}$, this avoids target leakage and ensures that evaluation-time performance reflects inference rather than supervised exposure.

\begin{table*}[t]
\centering
%\small
\setlength{\tabcolsep}{6pt}
\renewcommand{\arraystretch}{1.15}
\begin{tabular}{p{3.1cm} p{11.8cm}}
\toprule
\textbf{Stage} & \textbf{Example} \\
\midrule

Triples (Train Graph)
&
(1) (Edward Dane Burke, husband, Gabriel Bode Arnold) \newline
(2) (Gabriel Bode Arnold, wife, Edward Dane Burke) \newline
(3) (Gabriel Bode Arnold, friend, Michael Brett Ferguson) \newline
(4) (Michael Brett Ferguson, friend, Gabriel Bode Arnold) \newline
(5) (Edward Dane Burke, job, project manager) \newline
(6) (Gabriel Bode Arnold, job, consultant) \newline
(7) (Michael Brett Ferguson, job, entertainment manager) \\
\midrule

Simple paragraph (canonical verbalization)
&
(1) Edward Dane Burke is the husband of Gabriel Bode Arnold. \newline
(2) Gabriel Bode Arnold is the wife of Edward Dane Burke. \newline
(3) Gabriel Bode Arnold is the friend of Michael Brett Ferguson. \newline
(4) Michael Brett Ferguson is the friend of Gabriel Bode Arnold. \newline
(5) project manager is the job of Edward Dane Burke. \newline
(6) consultant is the job of Gabriel Bode Arnold. \newline
(7) entertainment manager is the job of Michael Brett Ferguson. \\
\midrule

Paragraph after template sampling + shuffle
&
(1) Edward Dane Burke serves as Gabriel Bode Arnold's husband. \newline
(2) Gabriel Bode Arnold serves as Michael Brett Ferguson's friend. \newline
(3) Michael Brett Ferguson acts in the role of friend to Gabriel Bode Arnold. \newline
(4) Edward Dane Burke is employed as a project manager. \newline
(5) Gabriel Bode Arnold is employed as a consultant. \newline
(6) Michael Brett Ferguson works as an entertainment manager. \newline
(7) Gabriel Bode Arnold is the wife of Edward Dane Burke. \\
\midrule

SFT corpus (QA generated from train content)
&
(1) Q: Who holds the relation of husband to Gabriel Bode Arnold? \newline
\hspace*{1.6em}A: Edward Dane Burke \newline
(2) Q: Who is the friend of Michael Brett Ferguson? \newline
\hspace*{1.6em}A: Gabriel Bode Arnold \newline
\ldots \newline
(6) Q: What does Michael Brett Ferguson work as? \newline
\hspace*{1.6em}A: entertainment manager \newline
\ldots \\
\bottomrule
\end{tabular}
\caption{End-to-end text generation pipeline: from triples to paragraph realizations, and to SFT QA pairs (generated only from the training content).}
\label{tab:text_pipeline_examples}
\end{table*}

\subsection{Other Results}

\subsubsection{Effect of increasing templates under fixed train samples}

\begin{figure*}[t]
\centering
\includegraphics[width=1\columnwidth]{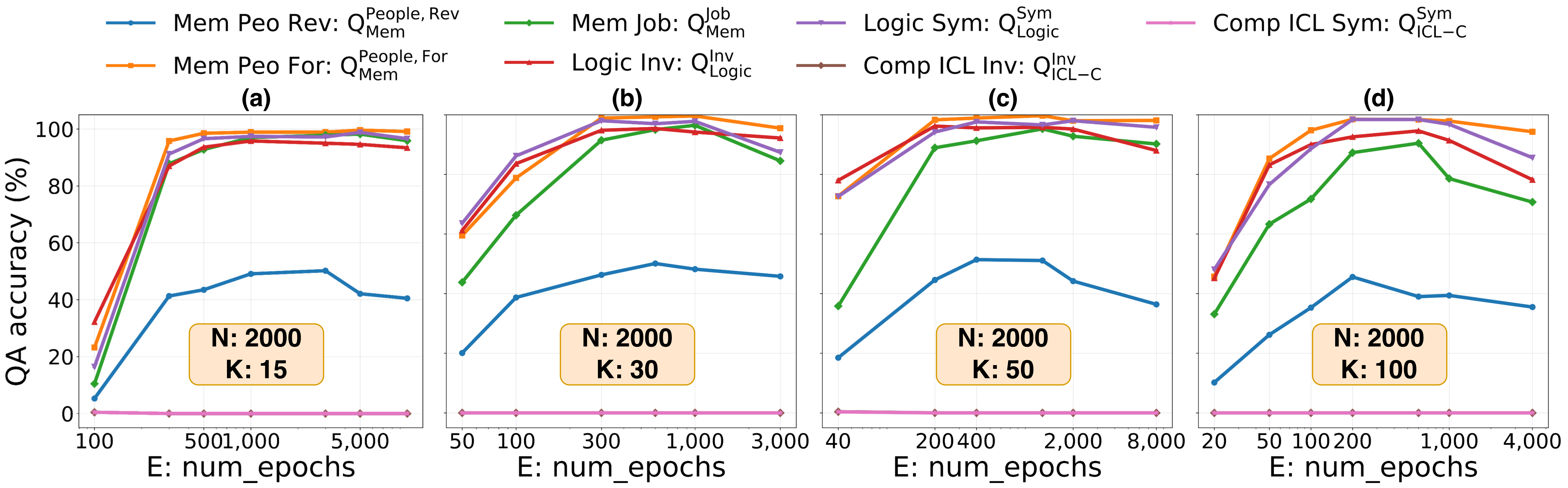}
\caption{Evaluation accuracy across training steps $E$ for $N=2000$, $L=12$, and template counts $K\in\{15,30,50,100\}$. Increasing $K$ alone does not induce in-context inversion/symmetry understanding.}

\label{fig:different_template}
\end{figure*}

With ($N=2000$) and ($L=12$) fixed, in addition to the ($K=10$) experiment that tracks evaluation accuracy as a function of training steps ($E$), we also ran experiments with ($K=15, 30, 50,$) and (100), corresponding to Figure~\ref{fig:different_template}(a)--(d). As shown in the figure, except for in-context Inversion and Symmetry which remain at 0 throughout, all other evaluation metrics increase at early stages and then decline as ($E$) grows. These results confirm the same training-step trend and further indicate that increasing only the number of templates ($K$) (without increasing the number of samples ($N$)) is insufficient to induce emergent logical understanding.

\subsubsection{More details and experiments on layer-wise analysis}

\paragraph{Experiment setup for layer-wise analysis for ICL completion} \label{sec:detail_layer_wise_analysis}
In Result~2 (Task~\ref{sec:layer_wise_analysis}), we analyze how model confidence evolves across layers by measuring, for \(N\in{2000, 10{,}000, 20{,}000}\), the logit, probability, and vocabulary rank of the \textbf{first correct answer token} conditioned on an in-context prompt. For a 12-layer GPT-2 model, the final next-token distribution is obtained by applying the shared output projection (LM head) to the final hidden state (after the last layer normalization), followed by softmax. To obtain layer-wise signals, for each layer \(l\) we take its intermediate hidden representation and decode it using the \textbf{same LM head} (and the corresponding normalization used for decoding), producing a full-vocabulary logit vector and the associated probabilities. We then extract the logit, probability, and rank of the gold answer’s first token (e.g., for the answer “Noah Dylan Martinez”, we track the token “Noah”). Repeating this procedure for all layers and all 1,800 prompts, we report per-layer mean logit, mean probability, and mean rank.

\begin{figure}[t]
\centering
\includegraphics[width=1\columnwidth]{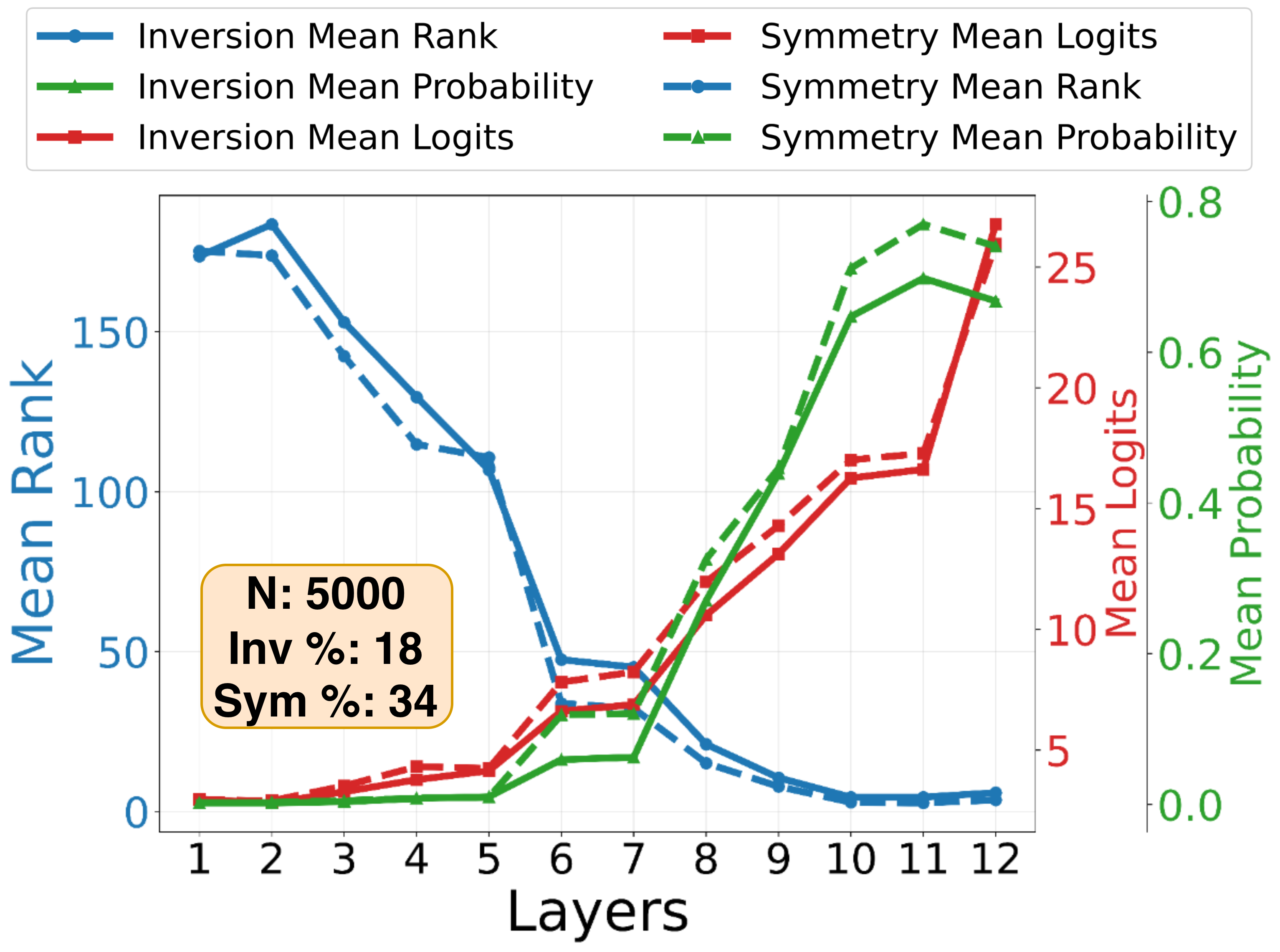}
\caption{Layer-wise mean logit, mean probability, and mean rank of the first correct token for the model trained with $N=5000$ and $K=10$, showing mid-layer emergence (layers 5--8) and stable late-layer improvement without last-layer collapse.}
\label{fig:logits_5000}
\end{figure}

\paragraph{More experiments about layer-wise analysis}
As shown in Figure~\ref{fig:logits_5000}, we additionally report the layer-wise mean logit, mean probability, and mean rank for the model trained on the synthetic corpus with \(N=5000\) and \(K=10\). Although its final accuracies are relatively low (Inversion: 18\%, Symmetry: 34\%), the overall layer-wise dynamics closely match those of the higher-performing models trained with \(N=10{,}000\) and \(N=20{,}000\): the model begins to improve rapidly in the mid layers (approximately layers 5--8) and continues to strengthen steadily in later layers, without the last-layer collapse observed in smaller-data regimes.

\begin{figure*}[t]
\centering
\includegraphics[width=1\columnwidth]{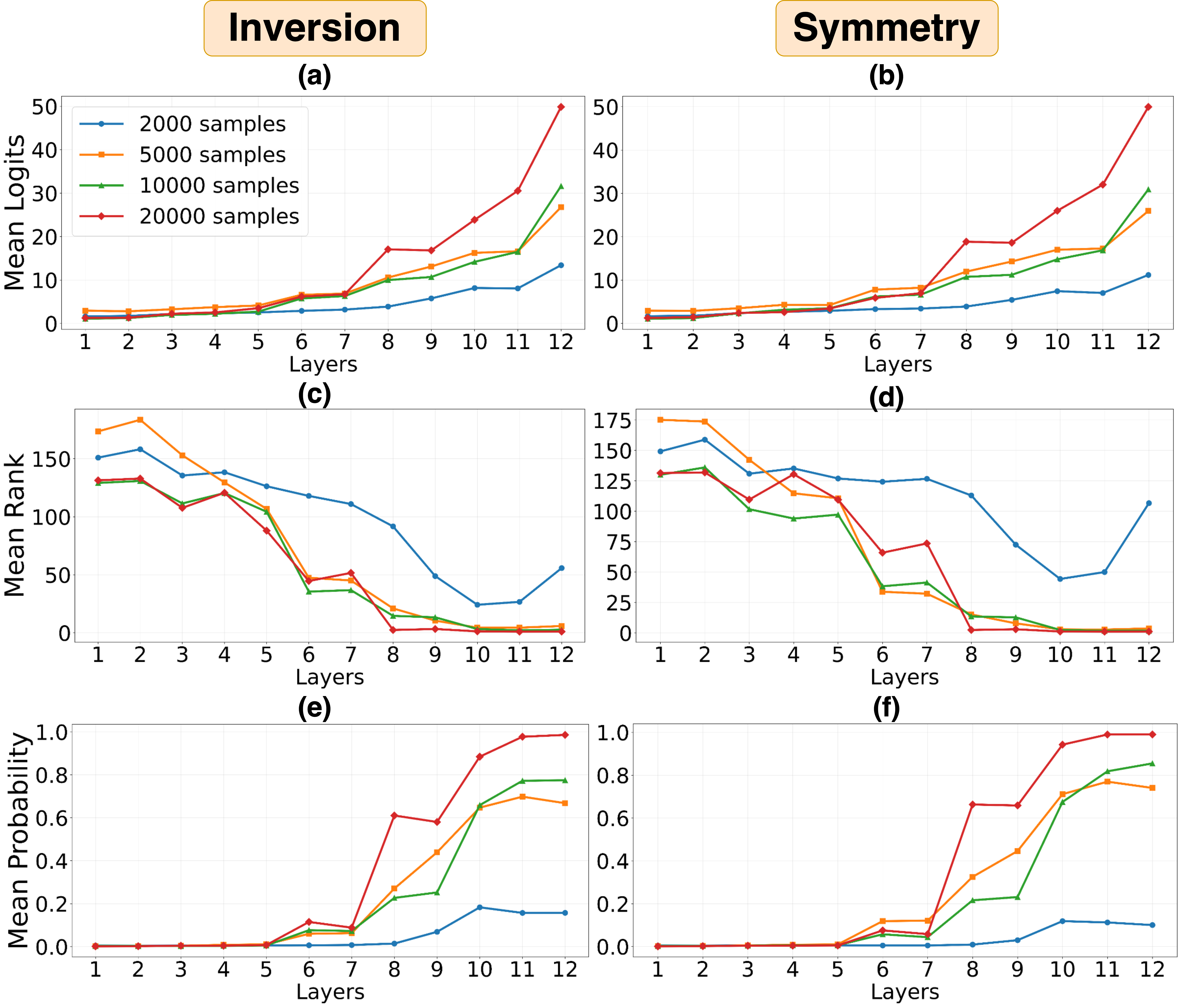}
\caption{Layer-wise mean logit, mean rank, and mean probability of the first correct token for $N\in\{2000,5000,10{,}000,20{,}000\}$ (Inversion and Symmetry shown separately), highlighting stronger and more stable late-layer behavior for larger $N$ and late-layer degradation for $N=2000$.}
\label{fig:logits_all_samples}
\end{figure*}

\begin{figure}[t]
\centering
\includegraphics[width=1\columnwidth]{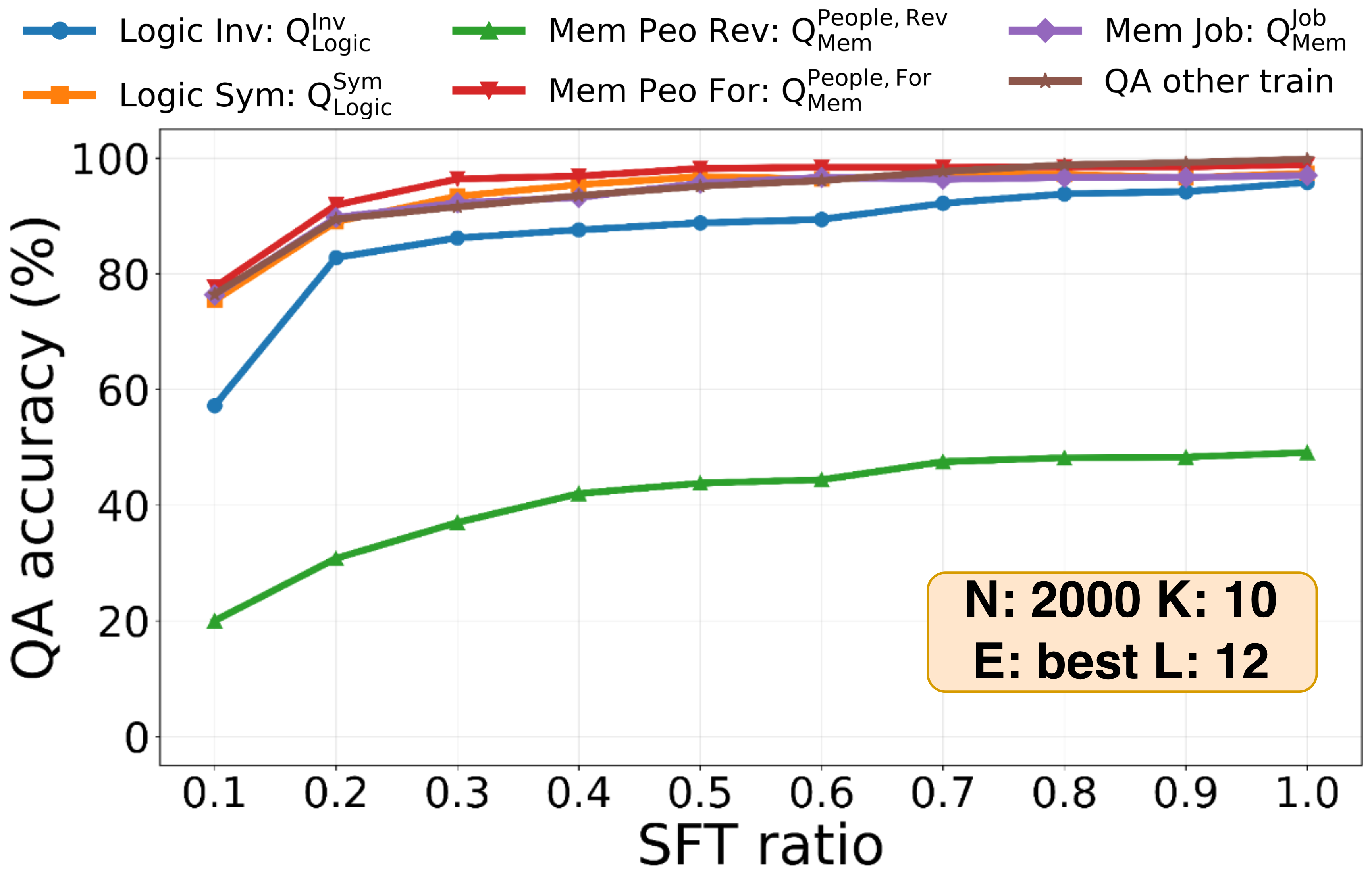}
\caption{Effect of SFT data fraction (10\%--100\%) on evaluation accuracy for the $N=2000$, $K=10$ pre-trained model, showing a sharp improvement around 20\% and strong performance with limited SFT data.}
\label{fig:sft_ratio}
\end{figure}

\begin{figure}[t]
\centering
\includegraphics[width=1\columnwidth]{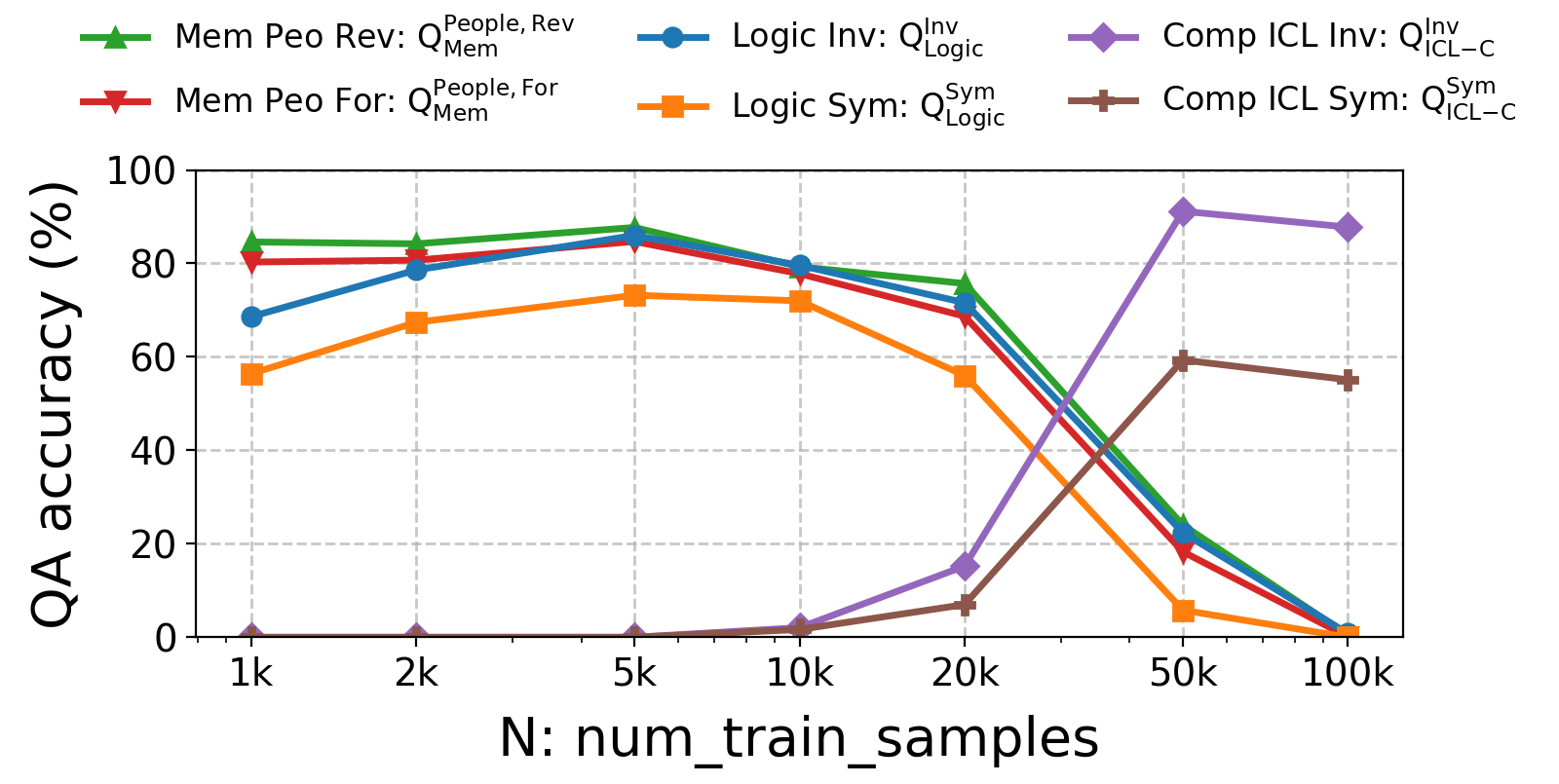}
\caption{Performance of the Diffusion Model across different training scales ($N$).}
\label{fig:diffusion_scale}
\end{figure}

To facilitate comparison, we aggregate the layer-wise mean logit, mean probability, and mean rank for \(N\in{2000, 5000, 10{,}000, 20{,}000}\) into a single figure, with Inversion and Symmetry reported separately.
Figure~\ref{fig:logits_all_samples} first shows in (a)(b) that the mean logits increase monotonically across layers for all \(N\), while larger \(N\) yields higher logits in the final layer; in particular, the \(N=20{,}000\) model attains a substantially higher final-layer logit than the other settings, especially compared to \(N=2000\). In (c)(d), the mean rank curves for \(N=5000, 10{,}000,\) and \(20{,}000\) drop sharply starting around layer 5 and reach strong performance by layer 8, continuing to improve through layer 12. By contrast, the \(N=2000\) model improves more slowly in mid layers, only dropping markedly around layer 8, and then exhibits a late-layer degradation with rank increasing at layers 11--12. The mean probability curves in (e)(f) mirror the rank trends, with the \(N=2000\) model maintaining very low probabilities overall. This consolidated view highlights the clear gap between the weak logit-level understanding under \(N=2000\) and the substantially stronger, stable layer-wise dynamics for \(N\ge 5000\), corroborating our earlier results and discussion.

\subsubsection{Effect of SFT data fraction}
To better understand the role of SFT, we conduct an additional experiment that varies the amount of SFT data. Our SFT corpus is derived from the Train Content: for each sentence verbalized from a triple, we construct a corresponding QA pair to equip the pre-trained model with question-answering capability. In all previous experiments, we fine-tuned on the full SFT corpus. Since the model pre-trained with \(N=2000\) and \(K=10\) already achieves strong QA performance after fine-tuning, we fix this pre-training setting and run SFT using only \(10\%\) to \(100\%\) of the SFT corpus. As shown in Figure~\ref{fig:sft_ratio}, we evaluate \(\text{Q}_{\text{Mem}}^{\text{People}}\) (forward and reverse), \(\text{Q}_{\text{Mem}}^{\text{Job}}\) (reverse), and \(\text{Q}_{\text{Logic}}\) (Inversion and Symmetry), and additionally test QA on training content excluded from SFT (e.g., when using 10\% of the SFT corpus, we sample QA instances from the remaining 90\% portion). We find that performance is modest at 10\%, but starting from 20\% the accuracies exhibit a clear turning point and already reach high values, indicating that even a small fraction of the SFT corpus is sufficient for effective fine-tuning.

\subsection{Experiments on diffusion models}
\label{app:diffusion_model}

\subsubsection{Implementation details}
To ensure a rigorous comparison with the AR baseline, our Diffusion LM is built upon the same \textbf{GPT-2 small} backbone ($L=12$, hidden size 768, 12 heads). However, we introduce key architectural adaptations to support the non-autoregressive training objective, strictly following the LLaDA framework \cite{nie2025large}:
\begin{itemize}
    \item \textbf{Bidirectional Attention}. We remove the causal masking in the self-attention layers. This allows every token to attend to the entire sequence context (both prefix and suffix) simultaneously, which is fundamental for learning joint probability distributions.
    \item \textbf{Training Objective}. The model is trained using a \textbf{Bernoulli masking} strategy. At each training step, tokens are independently masked with a probability $t \sim U(0,1)$, and the model optimizes the re-weighted cross-entropy loss on the masked positions.
    \item \textbf{Inference Strategy}. We employ a \textbf{block-wise generation} strategy with low-confidence re-masking. Specifically, the answer is generated in blocks of 4 tokens. Within each block, we iteratively refine the tokens by re-masking positions with the lowest confidence scores, allowing the model to self-correct based on the global bidirectional context.
\end{itemize}

\subsubsection{Data adjustment}
A critical challenge in applying diffusion models to relational logic is the potential for ``textual interpolation.'' In a standard paragraph containing coupled sentences, a uniform random mask rarely obscures one specific sentence entirely. Consequently, the model may learn to reconstruct missing tokens by interpolating from residual fragments of the symmetric sentence in the same context window, rather than learning the logical implication itself.

To address this, we restructured the pre-training corpus into three components with a ratio of \textbf{Raw : Single : Subset = 1 : 1 : 2}:
\begin{itemize}
    \item \textbf{Raw (25\%)}. Original complete paragraphs to maintain document-level coherence.
    \item \textbf{Single (25\%)}. Randomly sampled individual sentences. This forces the model to learn atomic syntax and entity structures without reliance on external context.
    \item \textbf{Subset (50\%)}. Random subsets containing 2 to $N-1$ sentences. This component is crucial as it simulates ``missing link'' scenarios, compelling the model to internalize the underlying logical rules to recover the missing information during training.
\end{itemize}

\subsubsection{Performance across different scales}
As shown in Figure~\ref{fig:diffusion_scale}, the diffusion model's In-context generalization improves with data scale, while rote memorization of Memorize QA facts fluctuates due to capacity constraints. Crucially, however, the immunity to the reversal curse persists across all scales, confirming that this robustness is intrinsic to the bidirectional architecture and holds regardless of data volume.

\subsubsection{Discussion: why does reverse outperform forward?}
Figure~\ref{fig:diffusion_scale} shows that reverse setting is on average about 4\% higher than the forward setting. We hypothesize this stems from the \textbf{structural proximity of entity anchoring} in our prompts. In Reverse queries (e.g., ``\textit{Who serves as [Person B]'s niece?}''), the target entity [Person B] is explicitly provided as a local anchor immediately adjacent to the answer slot, offering a clear, low-entropy guidance signal for the diffusion process.

\begin{table*}[t]
\centering
\small
\setlength{\tabcolsep}{6pt}
\renewcommand{\arraystretch}{1.15}
\begin{tabular}{p{5cm} p{10cm}}
\toprule
\textbf{Category} & \textbf{Templates} \\
\midrule

People--People sentence templates \(\mathcal{F}\) \newline
(person\_a, rel, person\_b)
&
(1) \{person\_a\} is the \{relationship\} of \{person\_b\}. \newline
(2) \{person\_a\} serves as \{person\_b\}'s \{relationship\}. \newline
(3) \{person\_a\} acts in the role of \{relationship\} to \{person\_b\}. \newline
(4) \{person\_a\} holds the relation of \{relationship\} to \{person\_b\}. \\
\midrule

People--Job sentence templates \(\mathcal{F}\) \newline
(person, Job, job)
&
(1) \{job\} is the job of \{person\}. \newline
(2) \{person\} works as \{article\} \{job\}. \newline
(3) \{person\}'s occupation is \{job\}. \newline
(4) \{person\} is employed as \{article\} \{job\}. \\
\midrule

People--People QA question templates (reverse) \newline
(person, relationship, ?)
&
(1) Who is the \{relationship\} of \{person\}? \newline
(2) Who serves as \{person\}'s \{relationship\}? \newline
(3) Who acts in the role of \{relationship\} to \{person\}? \newline
(4) Who holds the relation of \{relationship\} to \{person\}? \\
\midrule

People--People QA question templates (forward) \newline
(person, relationship, ?)
&
(1) \{person\} is the \{relationship\} of who? \newline
(2) \{person\} serves as whose \{relationship\}? \newline
(3) \{person\} acts in the role of \{relationship\} to who? \newline
(4) \{person\} holds the relation of \{relationship\} to who? \\
\midrule

People--Job QA question templates (reverse) \newline
(person, Job, ?)
&
(1) What is the job of \{person\}? \newline
(2) What does \{person\} work as? \newline
(3) What is \{person\}'s occupation? \newline
(4) What is \{person\} employed as? \\
\midrule

People--Job QA question templates (forward) \newline
(person, Job, ?)
&
(1) \{person\}'s job is what? \newline
(2) \{person\} works as what? \newline
(3) \{person\}'s occupation is what? \newline
(4) \{person\} is employed as what? \\
\midrule

In-context learning completion templates \newline
(person\_a, rel, person\_b)\newline
(person\_b, reverse\_rel, ?)
&
(1) \{person\_a\} serves as \{person\_b\}'s \{relationship\}, \{person\_b\} acts in the role of \{reverse\_relationship\} to \newline
(2) \{person\_a\} holds the relation of \{relationship\} to \{person\_b\}, \{person\_b\} is the \{reverse\_relationship\} of \newline
(3) \{person\_a\} acts in the role of \{relationship\} to \{person\_b\}, \{person\_b\} holds the relation of \{reverse\_relationship\} to \\
\bottomrule
\end{tabular}

\caption{Sentence, QA, and in-context completion templates used for synthetic data generation. For symmetric relations, \texttt{reverse\_relationship} is identical to \texttt{relationship}; for inversion relations, \texttt{reverse\_relationship} is the corresponding inverse relation. The in-context learning QA templates are the union of the People--People sentence templates and the People--Job QA reverse question templates.}

\label{tab:templates}
\end{table*}

\begin{table*}[t]
\centering
\small
\setlength{\tabcolsep}{6pt}
\renewcommand{\arraystretch}{1.15}
\begin{tabular}{p{3.2cm} p{11.8cm}}
\toprule

Evaluation content \newline
(paragraph example)
&
(1) Gregory Joel Henderson serves as Noah Dylan Martinez's husband. \newline
\textcolor{red}{\sout{(2) Noah Dylan Martinez is the wife of Gregory Joel Henderson.}} \newline
(3) Raymond Clayton Morris is the friend of Noah Dylan Martinez. \newline
\textcolor{red}{\sout{(4) Noah Dylan Martinez is the friend of Raymond Clayton Morris.}} \newline
(5) Gregory Joel Henderson is employed as a voice actor. \newline
(6) bus driver is the job of Noah Dylan Martinez. \newline
(7) Raymond Clayton Morris works as an event planner. \\
\midrule

\textbf{Evaluation} & \textbf{QA Example} \\
\midrule

Memorize QA (Reverse) \newline
$\text{Q}_{\text{Mem}}^{\text{People, Rev}}$ and $\text{Q}_{\text{Mem}}^{\text{Job, Rev}}$
&
People: Q: Who holds the relation of husband to Noah Dylan Martinez? \newline
\hspace*{2.3em}A: Gregory Joel Henderson \newline
Job: Q: What is the job of Gregory Joel Henderson? \newline
\hspace*{2.3em}A: voice actor \\
\midrule

Memorize QA (Forward) \newline
$\text{Q}_{\text{Mem}}^{\text{People, For}}$ and $\text{Q}_{\text{Mem}}^{\text{Job, For}}$
&
People: Q: Gregory Joel Henderson holds the relation of husband to who? \newline
\hspace*{2.3em}A: Noah Dylan Martinez \newline
Job: Q: Gregory Joel Henderson's job is what? \newline
\hspace*{2.3em}A: voice actor \\
\midrule

Logic QA (Forward)\newline
$\text{Q}_{\text{Logic}}^{\text{Inv, For}}$ and $\text{Q}_{\text{Logic}}^{\text{Sym, For}}$
&
Inversion: Q: Who holds the relation of wife to Gregory Joel Henderson? \newline
\hspace*{2.3em}A: Noah Dylan Martinez \newline
Symmetry: Q: Who is the friend of Raymond Clayton Morris? \newline
\hspace*{2.3em}A: Noah Dylan Martinez \\
\midrule

Logic QA (Reverse) \newline
$\text{Q}_{\text{Logic}}^{\text{Inv, Rev}}$ and $\text{Q}_{\text{Logic}}^{\text{Sym, Rev}}$
&
Inversion: Q: Noah Dylan Martinez holds the relation of wife to who? \newline
\hspace*{2.3em}A: Gregory Joel Henderson \newline
Symmetry: Q: Noah Dylan Martinez is the friend of who? \newline
\hspace*{2.3em}A: Raymond Clayton Morris \\
\midrule

ICL Logic QA \newline
(Completion)\newline
$\text{Q}_{\text{ICL-C}}^{\text{Inv}}$ and $\text{Q}_{\text{ICL-C}}^{\text{Sym}}$
&
Inversion: Q: Douglas Chance Holmes acts in the role of wife to Eric Braden Edwards.Eric Braden Edwards is the husband of \newline
\hspace*{2.3em}A: Douglas Chance Holmes \newline
Symmetry: Q: Alan Logan Butler holds the relation of spouse to Jeffrey Brent Brown. Jeffrey Brent Brown holds the relation of spouse to \newline
\hspace*{2.3em}A: Alan Logan Butler \\
\midrule

ICL Logic QA (QA) \newline
$\text{Q}_{\text{ICL-Q}}^{\text{Inv, For}}$ and $\text{Q}_{\text{ICL-Q}}^{\text{Sym, For}}$
&
Inversion: Q: Douglas Chance Holmes acts in the role of wife to Eric Braden Edwards. Who is the husband of Douglas Chance Holmes? \newline
\hspace*{2.3em}A: Eric Braden Edwards \newline
Symmetry: Q: Alan Logan Butler holds the relation of spouse to Jeffrey Brent Brown. Who holds the relation of spouse to Alan Logan Butler? \newline
\hspace*{2.3em}A: Jeffrey Brent Brown \\

\midrule

ICL Logic QA (QA) \newline
$\text{Q}_{\text{ICL-Q}}^{\text{Inv, Rev}}$ and $\text{Q}_{\text{ICL-Q}}^{\text{Sym, Rev}}$
&
Inversion: Q: Douglas Chance Holmes acts in the role of wife to Eric Braden Edwards. Eric Braden Edwards is the husband of who? \newline
\hspace*{2.3em}A: Douglas Chance Holmes \newline
Symmetry: Q: Alan Logan Butler holds the relation of spouse to Jeffrey Brent Brown. Jeffrey Brent Brown is the spouse of who? \newline
\hspace*{2.3em}A: Alan Logan Butler \\

\bottomrule
\end{tabular}
\caption{Examples of evaluation content and the corresponding evaluation question and answering.}
\label{tab:eval_icl_qa_examples}
\end{table*}

\end{document}